\documentclass[journal,10pt]{IEEEtran}
\IEEEoverridecommandlockouts

\usepackage{cite}
\usepackage{amsmath,amssymb,amsfonts}
\usepackage{graphicx}
\usepackage{textcomp}
\usepackage{xcolor}
\usepackage{booktabs}
\usepackage{array,tabularx}
\usepackage{multirow}
\usepackage{xurl}
\usepackage{stfloats}
\usepackage{balance}
\usepackage{xspace}
\usepackage{hyperref}
\usepackage{pdfcomment}
\usepackage{orcidlink}
\graphicspath{{Figures/}}

\def\BibTeX{{\rm B\kern-.05em{\sc i\kern-.025em b}\kern-.08em
    T\kern-.1667em\lower.7ex\hbox{E}\kern-.125emX}}

\newcommand{\qwk}{\ensuremath{\kappa_{w}}\xspace}

\begin{document}
\raggedbottom

\title{Vision-Language Models for Criterion-Level Grading of Handwritten Examinations in Outcome-Based Education}

\author{Asif~Hasan~Tonmoy,
        Saad~Ahmed\,\orcidlink{0009-0005-5614-3427},
        Md~Khalid~Syfullah\,\orcidlink{0009-0003-6219-5851},
        and~S.~M.~Jahangir~Alam
\thanks{This work was funded by the Bangladesh Accreditation Council (BAC)
under the project \emph{AI-Powered Automated Exam Evaluation and OBE-Based
University Assessment System}.}}

\markboth{IEEE Transactions on Learning Technologies}%
{Syfullah \MakeLowercase{\textit{et al.}}: Towards AI-Powered Automated Evaluation of Handwritten Exam Papers}

\maketitle

\begin{abstract}
Criterion-level grading connects examination performance to learning outcomes, but manual marking introduces workload and variation between markers. This study evaluates vision-language models (VLMs) for handwritten outcome-based assessment across five dimensions: accuracy, human agreement, repeated-run reliability, error concentration, and explanation quality. Using 1,982 criterion-level records from 485 undergraduate examination answers, we compare 20 configurations spanning Qwen2.5-VL, InternVL3, Pixtral, a Donut baseline, and a cascade ensemble. Evaluation setups include zero-shot prompting, few-shot prompting, partial fine-tuning, and Low-Rank Adaptation (LoRA). Two independent faculty markers regraded all 291 test criteria, providing a human agreement baseline on the same assessment materials. Qwen2.5-VL with LoRA achieved Quadratic Weighted Kappa (QWK) of 0.727 and mean absolute error of 0.435 marks against the examiner, compared with mean human-pair QWK of 0.551. This comparison reflects calibration to the examiner's training marks. LoRA outperformed partial fine-tuning for all three instruction-tuned VLMs, while few-shot prompting reduced QWK in every configuration with valid prompted scores. Aggregate reliability and exact repeatability diverged: intraclass correlations ranged from 0.790 to 0.874, yet 50.2--63.6\% of criteria changed marks across five sampled runs. Attention-guided deletion showed no statistically significant advantage over random masking, and four faculty reviewers reached no consensus on explanation usefulness. These findings highlight the need for rubric-specific calibration, repeatable scoring, review of consequential errors, and separate validation of explanations. The released evaluation protocol supports criterion-level assessment research and grading tools with teacher oversight.
\end{abstract}


\IEEEpeerreviewmaketitle

\section{Introduction}
\label{sec:introduction}


Outcome-Based Education (OBE) organizes teaching and assessment around explicit learning outcomes~\cite{spady1994outcome,harden2002learning,biggs2011teaching}. In criterion-based assessment, each awarded mark contributes evidence toward Course Learning Outcomes (CLOs) and Program Learning Outcomes (PLOs)~\cite{rajak2019copo,premalatha2019course,reddy2021copomapping}. Preserving this link requires repeated partial-credit decisions: eighty answers scored against four criteria produce 320 judgments. Studies of OBE adoption and marking practice identify the resulting workload and variation between markers as practical challenges~\cite{hassan2012obebangladesh,tisi2013review}. Automated criterion-level grading could support this assessment workflow by producing marks that remain attached to their rubric criteria. Establishing how accurately and consistently those marks can be generated is therefore an important learning-technology problem.


Two properties distinguish this task from much of the Automated Essay Scoring (AES) literature~\cite{page1968analysis,attali2006erater,shermis2013handbook,taghipour2016neural,ridley2021automated}. First, the target is a partial-credit mark for a specific criterion with a fixed maximum, rather than a holistic response score. Agreement on an answer total can conceal disagreement on its individual criteria. Criterion-level decisions are therefore needed when CLO attainment is computed from mapped criteria~\cite{rajak2019copo,ramchandra2014attainment}. Second, the input is a scanned handwritten answer rather than typed text. The model must interpret text, layout, equations, and figures together. A handwriting-recognition system followed by a text grader offers one approach~\cite{li2023trocr,crosilla2025htrllm}. However, a text-only transcription can omit diagrams, spatial relationships, or crossed-out work that may affect grading decisions~\cite{tisi2013review,jonsson2007rubrics}.


Vision-Language Models (VLMs) process page images with natural-language instructions~\cite{liu2023llava,bai2025qwen25vl,zhu2025internvl3,mistral2024pixtral}, while Parameter-Efficient Fine-Tuning (PEFT) supports adaptation with a modest training corpus~\cite{hu2021lora,dettmers2023qlora,mangrulkar2022peft}. These capabilities make rubric-guided handwritten grading feasible to investigate, but its educational usefulness depends on more than average accuracy. Williamson \textit{et al.}~\cite{williamson2012framework} emphasize agreement, reliability, and subgroup performance in automated scoring. Explainable learning analytics also requires decisions that educators can inspect~\cite{khosravi2022explainable}. We operationalize these requirements through five studies of 20 configurations: four backbones and a cascade ensemble under four setups. The central contribution is an integrated evaluation examining model adaptation, grading consistency, and explanation quality within the same criterion-level task. This connection is essential for selecting and designing assessment tools: a model may agree with the examiner overall while changing individual marks across runs or providing explanations that do not reflect its scoring behavior.


Our contributions are fourfold:
\begin{itemize}
\item \textbf{A criterion-level evaluation framework:} We organize 1,982 judgments from 485 handwritten answers across nine subjects, 12 question templates, and 47 rubric criteria~\cite{obedataset2026}. The benchmark includes Qwen2.5-VL-7B-Instruct~\cite{bai2025qwen25vl}, InternVL3-8B~\cite{zhu2025internvl3}, Pixtral-12B~\cite{mistral2024pixtral}, Donut-base~\cite{kim2022donut}, and a cascade, evaluated through prompting, LoRA~\cite{hu2021lora}, and partial or full fine-tuning. Shared scoring and agreement measures~\cite{cohen1968weighted,landis1977measurement} enable comparisons at the unit of outcome-based assessment.

\item \textbf{An empirical human reference:} We assign two additional faculty members to independently grade the test set and establish a human agreement baseline~\cite{williamson2012framework,jonsson2007rubrics}. This baseline distinguishes model agreement with the examiner whose marks were used for training from agreement among independent human markers.

\item \textbf{Evidence on consistency and review needs:} We conduct five stochastic runs for each LoRA-adapted backbone to assess repeated-measure reliability~\cite{williamson2012framework}. We also examine exact repeatability and errors by subject, answer content, and criterion weight to identify inconsistencies and review needs that aggregate agreement alone may obscure.

\item \textbf{Separate evaluation of explanation quality:} We evaluate score-conditioned attention maps~\cite{abnar2020attention,chefer2021transformer} through deletion tests against random masking~\cite{wu2024faithfulness} and faculty appraisal~\cite{khosravi2022explainable}. These assessments examine whether highlighted regions influence model scores and whether educators find explanations useful for reviewing marks.
\end{itemize}


We publicly release the full codebase, configurations, prompts, human-study materials, and results to support reuse of the evaluation and closer examination of the findings. Five research questions guide the analysis:

\begin{itemize}
\item\textbf{RQ1:} How accurately do VLMs award criterion-level marks to handwritten answers, and how does adaptation affect accuracy?
\item\textbf{RQ2:} How does model agreement with the examiner compare with agreement among human markers on the same criteria?
\item\textbf{RQ3:} How stable are marks when the same system repeatedly grades the same answer?
\item\textbf{RQ4:} How does error vary across subjects, answer content, and criterion weights?
\item\textbf{RQ5:} Do attention-based explanations reflect the model's scoring behavior and help faculty reviewers assess its decisions?
\end{itemize}


The remainder of this paper is organized as follows. Section~\ref{sec:background} introduces the background; Section~\ref{sec:related} reviews related work; Section~\ref{sec:method} describes the corpus and evaluation framework; Section~\ref{sec:results} reports the five studies; and Section~\ref{sec:conclusion} concludes. The appendix provides supplementary details to support the main text.
\section{Background}
\label{sec:background}


This section introduces preliminary concepts used in our work: outcome-based assessment, VLMs, model adaptation, and agreement measures.


\begin{figure}[t]
\centering
\includegraphics[width=0.96\columnwidth]{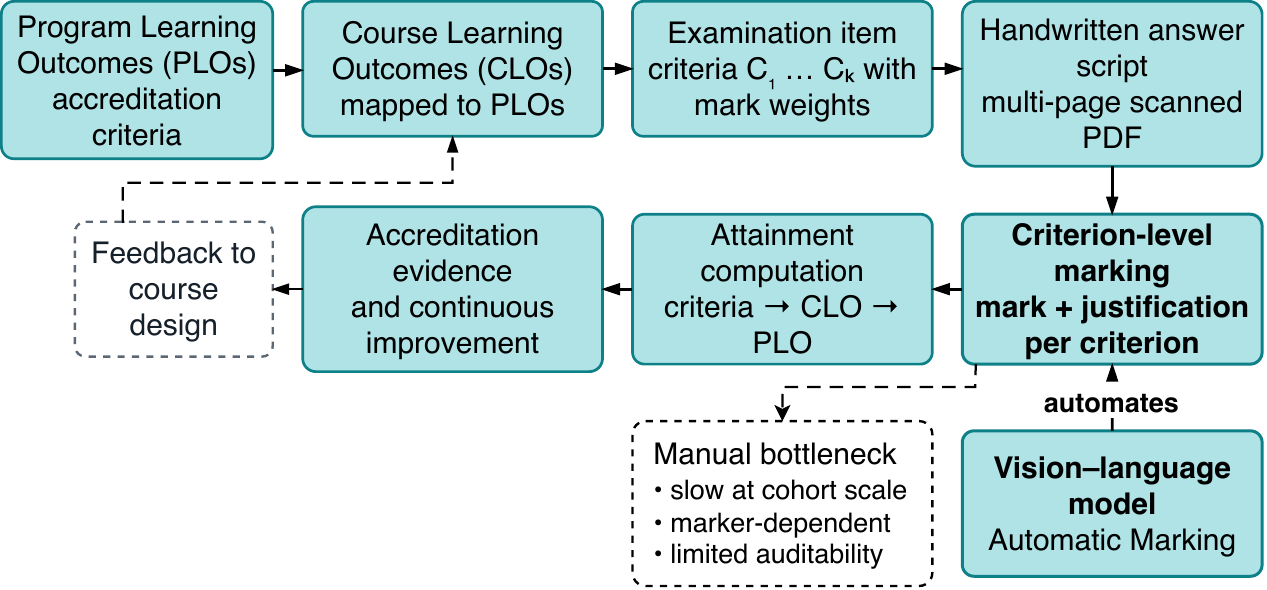}
\vspace{-10pt}
\caption{Assessment loop under outcome-based education. Marks flow from rubric criteria up to course and program outcomes, and the attainment report feeds back into course design. Criterion-level marking, shown in bold, is the manual step this work automates.}
\label{fig:obe}
\end{figure}


\subsection{Outcome-Based Assessment and Its Marking Step}

An OBE program defines what graduates should be able to do through PLOs. CLOs specify these expectations within individual courses, and examination items are mapped to relevant CLOs~\cite{spady1994outcome,harden2002learning}. An analytic rubric divides an item into named criteria, each with an allocated mark weight. Examiners score these criteria separately, and the resulting marks contribute to CLO and PLO attainment~\cite{rajak2019copo,premalatha2019course}. Fig.~\ref{fig:obe} shows this assessment loop and its feedback into course design. Automation must preserve the link between each mark and its criterion. Predicting an answer total alone cannot provide that criterion-level evidence.


Handwritten answers may contain diagrams, equations, crossed-out work, margin notes, and continuations on later pages. Differences in interpreting such material can contribute to marking variability~\cite{tisi2013review,jonsson2007rubrics}. A transcription-first pipeline may omit information that the rubric requires~\cite{li2023trocr,crosilla2025htrllm}. This possibility motivates direct image input, but we do not measure transcription loss or compare against an OCR-plus-grader pipeline. Donut provides a smaller OCR-free baseline; it does not isolate the effect of avoiding OCR.


\subsection{Vision-Language Models}

A VLM combines visual representations with language processing~\cite{radford2021clip,li2022blip,liu2023llava}. Generative VLMs encode image patches and use them with a text prompt to generate an answer. For grading, the prompt can contain the student image, question, reference answer, rubric criterion, and requested output format. We selected models with open weights, support for high-resolution images, and memory requirements compatible with LoRA on one accelerator. The selected instruction-tuned models were Qwen2.5-VL-7B-Instruct~\cite{bai2025qwen25vl}, InternVL3-8B~\cite{zhu2025internvl3}, and Pixtral-12B~\cite{mistral2024pixtral}. Under our memory budget, Pixtral used a 768 px long side, compared with 1280 px for the other two (Table~\ref{tab:hyper}). Model size and input resolution are therefore confounded. We also evaluated Donut-base~\cite{kim2022donut}, a smaller OCR-free document model with a visual encoder and autoregressive text decoder. Its architecture, scale, and pretraining differ from those of the instruction-tuned VLMs, so it is a practical baseline rather than a controlled test of language capability.


\subsection{Specializing a Model on a Small Corpus}

Updating every parameter of a billion-parameter model is costly, particularly with fewer than two thousand records. LoRA freezes pretrained weights and learns small low-rank updates~\cite{hu2021lora}. Restricting gradients and optimizer state to these updates reduces memory and can limit changes to pretrained capabilities~\cite{dettmers2023qlora,mangrulkar2022peft,biderman2024lora}. Unlike LoRA, partial fine-tuning updates selected decoder blocks while keeping the image encoder fixed. We compare these approaches to assess their accuracy under a single-accelerator budget.


\subsection{Judging Agreement on Marks}

Rubric marks form an ordered numerical scale. Exact-match accuracy treats a half-mark error and a three-mark error equally, so it cannot describe error severity. We therefore report chance-corrected agreement with distance-based weights~\cite{cohen1968weighted,landis1977measurement}, alongside errors in marks, Pearson and Spearman correlations, and parse coverage. Correlation measures association rather than agreement and does not replace error analysis. All measures are defined in Section~\ref{sec:method} and reported in Table~\ref{tab:metrics-all}.

\section{Related Work}
\label{sec:related}


We review four areas relevant to criterion-level grading: automated response scoring, handwriting and document understanding, VLM-based assessment, and OBE automation. Table~\ref{tab:related} summarizes their scope and remaining gaps.


\begin{table*}[!htb]
\vspace{-2pt}
\centering
\caption{Positioning against the research lines reviewed in Section~\ref{sec:related}. HW denotes direct reading of handwritten page images; VAL denotes validation against measured human agreement together with reliability and explanation evidence.}
\label{tab:related}
\vspace{-5pt}
\setlength{\tabcolsep}{4pt}
\footnotesize
\begin{tabular}{p{0.30\textwidth}p{0.13\textwidth}p{0.12\textwidth}cc p{0.24\textwidth}}
\hline
\textbf{Research line} & \textbf{Input} & \textbf{Prediction} & \textbf{HW} & \textbf{VAL} & \textbf{Remaining gap} \\ \hline
Automated essay scoring~\cite{page1968analysis,attali2006erater,taghipour2016neural,ridley2021automated} & Typed essays & Holistic score & No & Partial & Typed input; no criterion-level credit \\
Short-answer grading~\cite{dong2016automatic,cozma2018essay} & Typed short text & Item score & No & Partial & Short typed spans only \\
Handwriting recognition~\cite{li2023trocr,crosilla2025htrllm} & Handwritten images & Transcript & Yes & No & Stops at transcription quality \\
Document understanding~\cite{kim2022donut,ma2024mmlongbench,mondal2024icdarhwd} & Page images & Answer or grade & Yes & No & Holistic grade, no rubric criteria \\
VLM and LLM grading~\cite{rodrigues2025gpt4fair,ramesh2022autonomous,wu2024faithfulness} & Images and text & Grade with text & Yes & No & No human ceiling, no faithfulness test on marks \\
OBE automation~\cite{rajak2019copo,ramchandra2014attainment,reddy2021copomapping} & Recorded marks & Attainment report & No & No & Automates bookkeeping, not marking \\
\textbf{This work} & Handwritten images & Criterion mark & Yes & Yes & 5 studies against a measured ceiling \\
\hline
\end{tabular}
\end{table*}


\subsection{Automated Scoring of Written Responses}

Early Automated Essay Scoring (AES) systems used surface features and regression to predict grades~\cite{page1968analysis,attali2006erater,shermis2003automated}. Later systems learned representations end-to-end using neural models and evaluated both overall scores and essay traits~\cite{taghipour2016neural,dong2016automatic,cozma2018essay,ridley2021automated}. Reviews report that automated scores on some essay benchmarks can approach human inter-rater agreement~\cite{shermis2013handbook,ramesh2022autonomous}. A review of 75 rubric studies also associates improved consistency with analytic, topic-specific rubrics~\cite{jonsson2007rubrics}. These findings support human agreement as a reference for evaluation. However, they do not directly establish performance on handwritten examination images scored against individual partial-credit criteria.


\subsection{Handwriting Recognition and Document Understanding}

Transformer models have advanced handwritten text recognition~\cite{li2023trocr,crosilla2025htrllm}, while OCR-free document models process page images for structured prediction and question answering~\cite{kim2022donut,ma2024mmlongbench}. Handwritten-document benchmarks now include page-level recognition and visual question answering~\cite{mondal2024icdarhwd}. These tasks establish document-reading capabilities, but their targets are transcriptions or answers, rather than examiner-awarded marks for rubric criteria. Their results do not directly measure partial-credit grading agreement.


\subsection{Vision-Language Models in Assessment}

Visual instruction tuning, document question answering, and attribution analysis provide foundations for image-based assessment~\cite{liu2023llava,ma2024mmlongbench,wu2024faithfulness}. Recent work has also examined large language models for short-answer grading, including leniency and demographic bias~\cite{rodrigues2025gpt4fair}. A systematic review describes the methods and limitations of the broader AES field~\cite{ramesh2022autonomous}. Research on explainable learning analytics highlights the need for inspectable assessment decisions~\cite{khosravi2022explainable}. Separately, studies of vision-transformer explanations question whether attribution methods consistently outperform random controls~\cite{wu2024faithfulness}. These research areas motivate joint evaluation of scoring and explanations, but do not establish their combined validity for criterion-level marks on handwritten answers.


\subsection{Automation Within Outcome-Based Education}

OBE automation has largely focused on processes after marking: mapping course and program outcomes, computing attainment, and preparing accreditation reports~\cite{rajak2019copo,ramchandra2014attainment,reddy2021copomapping,premalatha2019course}. Adoption studies identify marking workload and inconsistency as continuing implementation challenges~\cite{hassan2012obebangladesh}, consistent with the broader marking literature~\cite{tisi2013review,jonsson2007rubrics}. Automating criterion-level scoring could therefore complement existing attainment-reporting tools.


\subsection{Research Gap and Positioning}

The reviewed literature establishes complementary capabilities in response scoring, document understanding, and assessment reporting. Our contribution connects these capabilities at the rubric criterion, where a handwritten response becomes evidence for a learning outcome. The framework in Table~\ref{tab:related} evaluates accuracy, human agreement, repeatability, error concentration, and explanations within one assessment setting. Their joint analysis provides design knowledge that a model ranking alone cannot supply: which adaptations improve calibration, whether stable aggregate agreement yields repeatable individual marks, and whether accompanying explanations support review.

\section{Corpus and Evaluation Framework}
\label{sec:method}


\begin{table}[!t]
\vspace{-2pt}
\centering
\caption{Summary of notation used throughout the paper.}
\label{tab:notation}
\vspace{-5pt}
\setlength{\tabcolsep}{4pt}
\scriptsize
\begin{tabular}{|c|l|}
\hline
\textbf{Symbol} & \textbf{Description} \\ \hline
$x,\ q$ & Stitched answer image; question with ref. answer \\ \hline
$C_j,\ D_j$ & Target criterion $j$ and its rubric level descriptors \\ \hline
$m_j$ & Mark weight (maximum) of criterion $j$, 0.5 to 4 \\ \hline
$s_j,\ \hat{s}_j$ & Gold and predicted mark for criterion $j$ \\ \hline
$n$ & No. of scored judgments ($n=291$ at test) \\ \hline
$N$ & No. of attempted records, $N \geq n$ \\ \hline
$\delta$ & Mark grid resolution ($\delta=0.05$ marks) \\ \hline
$a_i,\ b_i$ & Gold and predicted grid positions of item $i$ \\ \hline
$L$ & No. of grid levels in a comparison ($L=81$) \\ \hline
$\kappa_w$ & Quadratic weighted kappa (QWK), Eq.~\eqref{eq:qwk} \\ \hline
$\bar{\kappa}_{H}$ & Mean pairwise QWK among the three human sources \\ \hline
MAE, RMSE & Mean absolute and root mean squared error in marks \\ \hline
nMAE & MAE normalized by the criterion mark weight $m_j$ \\ \hline
EM, AM & Exact and within-one-mark match rates, Eq.~\eqref{eq:match} \\ \hline
$r_{P},\ r_{S}$ & Pearson and Spearman correlation, $s_i$ vs.\ $\hat{s}_i$ \\ \hline
PR & Parse rate: share yielding a mark, Eq.~\eqref{eq:pr} \\ \hline
$W,\ \Delta W$ & Pretrained weight and its low-rank update $BA$ \\ \hline
$d_{\text{in}},d_{\text{out}}$ & Input and output dimensions of $W$ \\ \hline
$r,\ \alpha$ & LoRA rank and scaling factor ($r=16$, $\alpha=32$) \\ \hline
$k$ & Few-shot exemplars per backbone ($k=1$ or $3$) \\ \hline
$R$ & Repeated stochastic runs per system ($R=5$) \\ \hline
$\text{MS}_{I},\text{MS}_{R},\text{MS}_{E}$ & Item, run, residual mean squares, Eq.~\eqref{eq:icc} \\ \hline
ICC(2,1) & Two-way random-effects single-measure reliability \\ \hline
$\rho,\ \Delta_{\rho}$ & Masked patch fraction and resulting mark change \\ \hline
\end{tabular}
\end{table}


This section describes corpus preparation, prompting and output recovery, learning setups, the cascade ensemble, and the study protocols. The framework maps each scanned answer to a criterion-level mark and evaluates that mark from five complementary perspectives. Table~\ref{tab:notation} summarizes the notation.


\begin{figure*}[tp]
\centering
\includegraphics[width=0.92\textwidth]{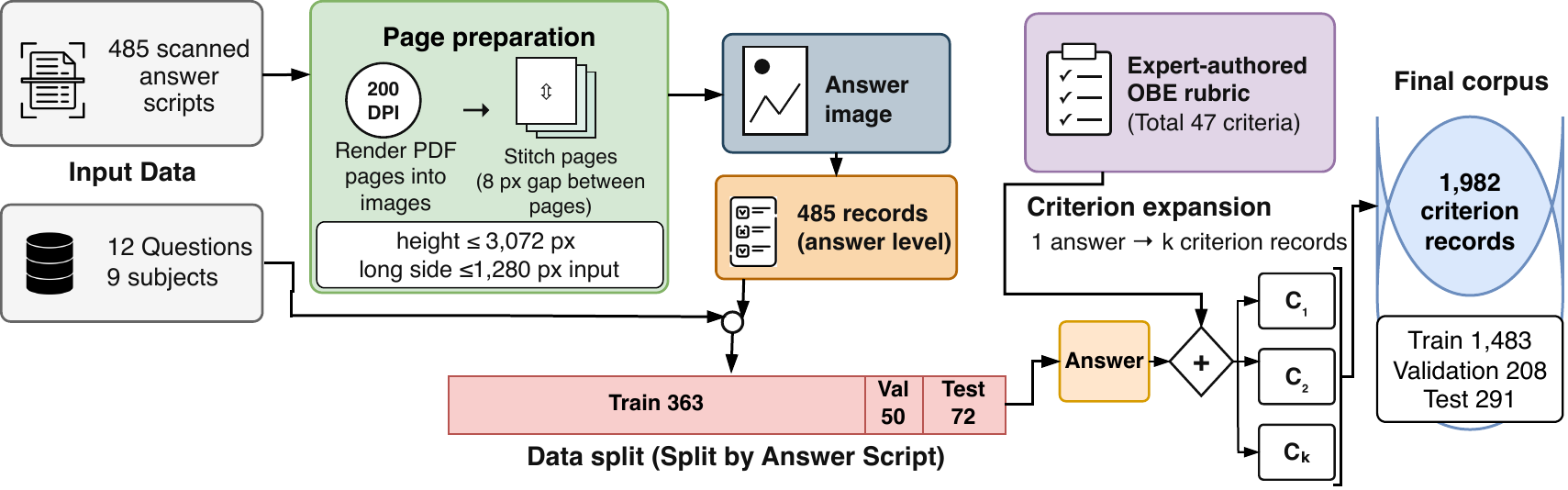}
\vspace{-10pt}
\caption{Corpus preparation process. Scanned scripts are rendered and stitched into one image per answer, split at the answer-script.}
\label{fig:dataset}
\end{figure*}


\subsection{Corpus and Preparation}
\label{sub:corpus}

The corpus contains 485 scanned final-examination answers from undergraduate students at four institutions and is released separately~\cite{obedataset2026}. Its twelve question templates cover nine subjects (Table~\ref{tab:dataset}). Eleven templates appear in the test split as Q1--Q11 (Table~\ref{tab:mapping}); five answers to the e-commerce template are in the training split. Seven templates elicit prose. Two Database Systems templates require tables and equations, Computer Networks includes an optional layered diagram, Object Oriented Programming includes code, and Algorithms includes equations and graph diagrams. Rubrics developed with an OBE expert decomposed the templates into 47 criteria, with maximum marks from 0.5 to 4. The course examiner assigned the reference, or gold, marks at the criterion level.


Fig.~\ref{fig:dataset} summarizes preparation. Multi-page PDFs were rendered at 200 dots per inch and stitched vertically with an 8 px gap into one image per answer. Images were capped at 3072 px in height and resized for each model as specified in Table~\ref{tab:hyper}. These limits were fixed before evaluation to satisfy memory and context constraints. Splitting at the answer-script level produced 363 training, 50 validation, and 72 test answers, corresponding to 1,483, 208, and 291 criterion-level records. A leakage check found no script in multiple splits. Criterion maxima were obtained from rubric metadata through a criterion-identity lookup built on the training split, replacing inconsistent maxima in the raw export. Gold and predicted marks were clamped to these maxima. To retain fractional marks such as 0.5, 1.25, and 1.8 without rounding to whole marks, kappa metrics used the 0.05-mark grid $\delta$, the coarsest grid representing the observed values exactly. The inclusive range from 0 to 4 contains $L=4/0.05+1=81$ levels. Retaining unused intermediate levels preserves numerical distances; their presence alone does not make weighted kappa conservative. Agreement remains sensitive to the score distributions and should be interpreted alongside error measures.


\begin{table}[!t]
\vspace{-2pt}
\centering
\caption{Composition of the corpus. Q, C and A are question templates, rubric criteria and answers; Score is the mean awarded mark divided by the rubric maximum of the question; $n$ is the number of criterion-level judgments contributed to the test split.}
\label{tab:dataset}
\vspace{-5pt}
\setlength{\tabcolsep}{2pt}
\scriptsize
\begin{tabular}{lrrrlrr}
\hline
\textbf{Subject} & \textbf{Q} & \textbf{C} & \textbf{A} & \textbf{Answer content} & \textbf{Score} & \textbf{$n$} \\ \hline
Machine Learning & 2 & 8 & 88 & Text & 0.78 & 44 \\
Digital Image Processing & 2 & 8 & 75 & Text & 0.77 & 28 \\
Database Systems & 2 & 8 & 80 & Text + tables + eq. & 0.63 & 64 \\
Computer Networks & 1 & 5 & 70 & Text + optional diagrams & 0.65 & 45 \\
Data Mining & 1 & 4 & 42 & Text & 0.56 & 32 \\
Algorithms & 1 & 4 & 83 & Text + eq. + diagrams & 0.83 & 40 \\
Fisheries & 1 & 3 & 23 & Text & 0.92 & 18 \\
Object Oriented Prog. & 1 & 4 & 19 & Text + code & 0.54 & 20 \\
E-commerce & 1 & 3 & 5 & Text & 0.49 & 0 \\
\hline
\textbf{Total} & \textbf{12} & \textbf{47} & \textbf{485} &  & & \textbf{291} \\ \hline
\end{tabular}
\end{table}


\begin{table*}[!htb]
\centering
\caption{Rubric criteria of each question template. Content codes are Tab for tables, Eq for equations, Diag for diagrams and Code for program code. C is the number of criteria the template defines and $n$ the number of criterion-level judgments it contributes to the test split.}
\label{tab:mapping}
\vspace{-6pt}
\setlength{\tabcolsep}{3pt}
\scriptsize
\begin{tabular}{p{0.1\textwidth}p{0.08\textwidth}ccp{0.68\textwidth}r}
\hline
\textbf{Subject} & \textbf{Content} & \textbf{Q} & \textbf{C} & \textbf{Criteria} & \textbf{n} \\ \hline
\multirow{2}{\linewidth}{Machine Learning} & Text & Q1 & 4 & importance of $R^2$ and variability; mathematical expression and interpretation; critical insight (limitations, adjusted $R^2$); enables comparison between models and overfitting & 8 \\
 & Text & Q2 & 4 & supervised learning; unsupervised learning; reinforcement learning; example, overall clarity and structure & 36 \\ \hline
\multirow{2}{\linewidth}{Digital Image Processing} & Text & Q3 & 4 & definition of quantization; purpose of quantization; Max--Lloyd algorithm explanation; boundary and centroid conditions & 12 \\
 & Text & Q4 & 4 & concept of histogram equalization; necessity of histogram equalization; concept of histogram matching; necessity of histogram matching & 16 \\ \hline
\multirow{2}{\linewidth}{Database Systems} & Text+Tab+Eq & Q5 & 4 & understanding of normalization and related problems; explanation of normal forms; use of example; clear and organized writing & 40 \\
 & Text+Tab+Eq & Q6 & 4 & understanding of schema and instance; explanation of key types with examples; correct identification of keys in a given relation; clear and organized writing & 24 \\ \hline
Computer Networks & Text+Diag & Q7 & 5 & identification of layers; explanation of layers with diagram; technical accuracy; examples and protocols; clarity and organization & 45 \\ \hline
Data Mining & Text & Q8 & 4 & definition of data mining; KDD process steps; importance of data mining; example and application & 32 \\ \hline
Algorithms & Text+Eq+Diag & Q9 & 4 & understanding of concept; step-by-step procedure; accuracy of final answer; clarity and presentation & 40 \\ \hline
Fisheries & Text & Q10 & 3 & definition of climate and weather; names of the climate variables; hemato-biochemical changes in rohu under thermal stress & 18 \\ \hline
Object Oriented Prog. & Text+Code & Q11 & 4 & definition of method overloading; definition of method overriding; comparison; example and application & 20 \\ \hline
\multicolumn{3}{l}{\textbf{Total (test split)}} & \textbf{44} & & \textbf{291} \\ \hline
\end{tabular}
\end{table*}


\subsection{Task Definition and Output Contract}

\begin{figure*}[tp]
\centering
\includegraphics[width=0.92\textwidth]{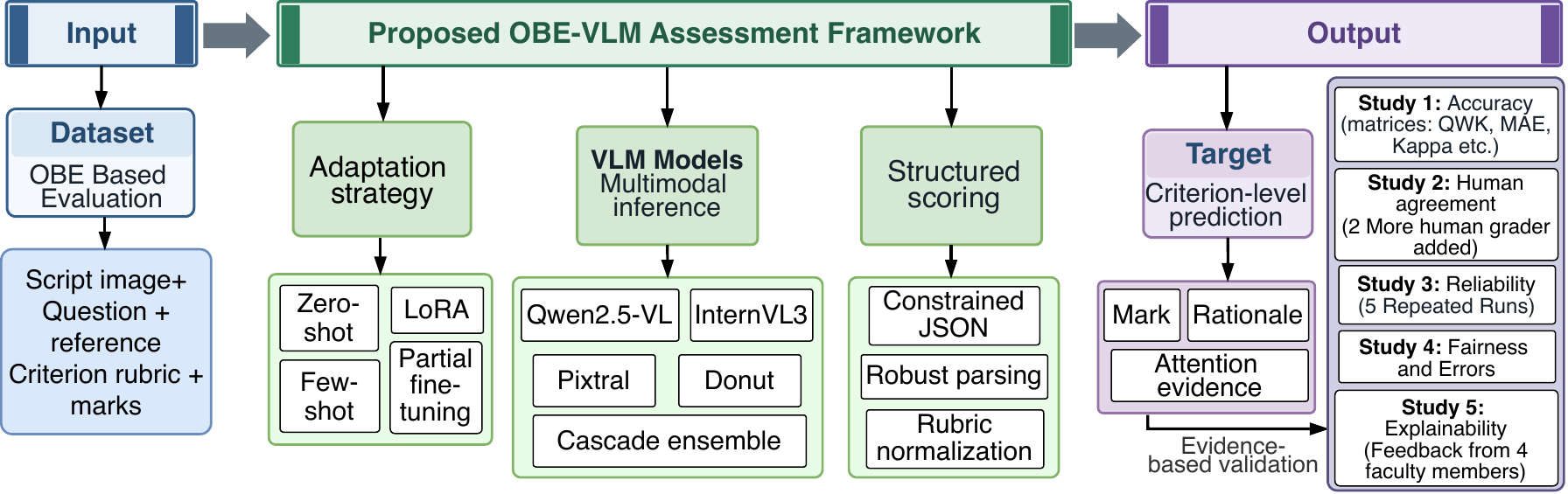}
\vspace{-10pt}
\caption{Evaluation framework. A criterion-level record is rendered into a chat prompt with an explicit output contract, graded by one of 20 systems formed from four backbones and a cascade ensemble under four learning setups, decoded into a bounded mark with a rationale, and evaluated by five studies.}
\label{fig:framework}
\end{figure*}

Let $x$ denote the stitched answer image, $q$ the question with its reference answer, and $C_j$ the target criterion with rubric-level descriptors $D_j$ and mark weight $m_j$. A grader is a function
\begin{equation}
f: (x, q, C_j, D_j, m_j) \mapsto \hat{s}_j \in [0, m_j],
\label{eq:task}
\end{equation}
where the gold mark $s_j$ was awarded by the course examiner. Fig.~\ref{fig:framework} shows how the components fit together. The same prompt, parser, clamp, and metric code serve every backbone and every learning setup, providing a $5\times4$ benchmark grid. Architecture-specific preprocessing and training differences limit controlled comparisons.


Each record is presented as a two-turn chat prompt. The system turn specifies the grading role and output format; the user turn contains the answer image, question, reference answer, target criterion, level descriptors, and maximum mark. The required output is one single-line JavaScript Object Notation (JSON) object, \verb|{"score": s, "justification": t}|. Where a chat template does not support a system role, its instructions are included in the user turn. The released templates document these adaptations. A recovery parser extracts JSON, handles code fences and truncated tails, and uses the first numeric literal only if no object can be recovered. This numeric fallback occurred only for Donut. Strict JSON parsing failed on 155/291 Qwen2.5-VL and 28/291 InternVL3 zero-shot outputs, 38 and 5 of their few-shot outputs, and 7 InternVL3 partial-fine-tuning outputs. Pixtral outputs and adapted Qwen2.5-VL outputs parsed without repair. Thus, full numeric parse coverage does not imply strict format compliance. Recovered marks are clamped to $[0,m_j]$ and snapped to $\delta$. Outputs without a numeric mark are counted as parse failures and excluded from score-based metrics. The parse rate $\text{PR}$ reports coverage; agreement on incomplete coverage is conditional on the records that parsed, and may be affected by selection bias.


\subsection{Learning Setups and the Cascade Ensemble}

Four setups are evaluated for each backbone. \textit{Zero-shot} supplies the assessment materials without examples. \textit{Few-shot} adds $k$ solved training examples, including their images, stratified over the normalized score range. The largest exemplar count that fit the context budget for all test items was used: $k=3$ for Qwen2.5-VL and Pixtral and $k=1$ for InternVL3. These counts were not selected by validation agreement, so few-shot comparisons do not hold $k$ constant. Donut does not support in-context examples; its few-shot cell uses the zero-shot prompt as a baseline. \textit{LoRA} freezes the pretrained weight matrix $W\in\mathbb{R}^{d_{\text{out}}\times d_{\text{in}}}$ and learns a low-rank update $\Delta W=BA$, with $r\ll\min(d_{\text{in}},d_{\text{out}})$:

\begin{equation}
W' = W + \tfrac{\alpha}{r} BA,
\label{eq:lora}
\end{equation}
with scaling factor $\alpha$~\cite{hu2021lora}. Adapters are inserted into attention and feed-forward projections on the language side while the vision encoder remains frozen. \textit{Partial fine-tuning} updates a suffix of decoder blocks chosen to approximate 30\% of the model parameters. The vision encoder, multimodal projector, and token embeddings remain frozen; Table~\ref{tab:hyper} lists the selected blocks. Donut's fourth configuration uses full fine-tuning of its encoder-decoder architecture. It occupies the same grid position for reporting, but is not equivalent to the partial fine-tuning used for the three VLMs.


The cascade first obtains marks from Qwen2.5-VL and InternVL3. If they differ by at most one mark, it returns their mean. Otherwise, it invokes Pixtral and averages all three marks. The threshold matches the absolute tolerance in Eq.~\eqref{eq:match}. Neither the threshold nor the aggregation rule was tuned. The result is clamped to the criterion maximum and snapped to a 0.25-mark grid, reflecting the commonly used quarter-mark step in this corpus. Because some reference marks lie between quarter points, this coarser grid restricts the cascade's possible outputs and affects comparisons with systems scored on $\delta=0.05$.


\subsection{Metrics}

Let $s_i$ and $\hat{s}_i$ be the gold and predicted marks of item $i$, and let $a_i$ and $b_i$ be their integer positions on the grid $\delta$ over $L$ levels. \textit{Quadratic weighted kappa} is the primary metric, comparing the observed agreement matrix $O$ against the chance matrix $E$ formed from the marginals and weighting each cell by squared grid distance:
\begin{equation}
\qwk = 1 - \frac{\sum_{u,v} w_{uv} O_{uv}}{\sum_{u,v} w_{uv} E_{uv}}, \qquad w_{uv} = \frac{(u-v)^2}{(L-1)^2}.
\label{eq:qwk}
\end{equation}
Linear weighted kappa replaces $w_{uv}$ with $|u-v|/(L-1)$, and unweighted kappa sets $w_{uv} = \mathbb{1}[u \neq v]$. \textit{Error magnitude} is reported as Mean Absolute Error (MAE), Root Mean Squared Error (RMSE), and signed bias:
\begin{equation}
\text{MAE} = \frac{1}{n}\sum_{i=1}^{n} |\hat{s}_i - s_i|, \quad \text{RMSE} = \sqrt{\frac{1}{n}\sum_{i=1}^{n} (\hat{s}_i - s_i)^2},
\label{eq:mae}
\end{equation}
\begin{equation}
\text{Bias} = \frac{1}{n}\sum_{i=1}^{n} (\hat{s}_i - s_i),
\label{eq:bias}
\end{equation}
together with the weight-normalized error $\text{nMAE} = \frac{1}{n}\sum_i |\hat{s}_i - s_i| / m_i$, which places every criterion on a common scale. \textit{Match rates} give the share of items graded exactly and within one mark:
\begin{equation}
\text{EM} = \frac{1}{n}\sum_{i=1}^{n} \mathbb{1}[a_i = b_i], \quad \text{AM} = \frac{1}{n}\sum_{i=1}^{n} \mathbb{1}[|\hat{s}_i - s_i| \leq 1].
\label{eq:match}
\end{equation}
AM applies a fixed one-mark tolerance and is therefore uninformative on criteria whose weight is one mark or less, where it is satisfied by construction; it is read beside EM and the weight-normalized nMAE for that reason rather than on its own.


Pearson correlation $r_P$ measures linear association between raw marks, and Spearman correlation $r_S$ measures rank association. Neither is chance-corrected; both can be affected by restricted score ranges and do not establish agreement. Parse coverage is
\begin{equation}
\text{PR} = \frac{1}{N}\sum_{i=1}^{N} \mathbb{1}[\text{output } i \text{ yields a mark}],
\label{eq:pr}
\end{equation}
over all $N$ attempted records, so $n \leq N$ for every other metric. \textit{Reliability} across $R$ repeated runs uses the two-way random-effects, single-measure intraclass correlation:
\begin{equation}
\text{ICC(2,1)} = \frac{\text{MS}_{I} - \text{MS}_{E}} {\text{MS}_{I} + (R-1)\text{MS}_{E} + \frac{R}{n}(\text{MS}_{R} - \text{MS}_{E})},
\label{eq:icc}
\end{equation}
where $\text{MS}_{I}$, $\text{MS}_{R}$, and $\text{MS}_{E}$ are the item, run, and residual mean squares. We also report mean per-item standard deviation and the fraction of items scored identically in all runs. \textit{Uncertainty} for pairwise human agreement in Study 2 is estimated using $B=2{,}000$ item-level bootstrap resamples and percentile 95\% intervals (Table~\ref{tab:human}). We compare model QWK point estimates with the largest human-pair upper bound descriptively. This is not a significance test of the model--human difference: model uncertainty is not estimated, and resampling criteria individually does not account for dependence among criteria from the same answer.


\subsection{The Five Studies and the Collection Instrument}

\textbf{Study 1 (accuracy)} evaluates all 20 configurations on the 291 test criteria. \textbf{Study 2 (human agreement)} establishes a measured human baseline. Two additional faculty markers independently graded all 291 criteria from the 72 test answers using a browser booklet. For each answer, it displayed the stitched image, question, reference answer, and criterion-level rubric descriptors and maxima. It concealed model outputs, examiner marks, and the other marker's scores. The assessment materials therefore matched those provided to the models, although the adapted models had also learned from the examiner's training marks. We report mean pairwise human QWK, $\bar{\kappa}_H$, and compare model point estimates with the human bootstrap bounds as described above.


\textbf{Study 3 (reliability)} repeats grading $R=5$ times for each LoRA-adapted backbone, varying decoding seeds at temperature 0.7. This probes sampling variability, not training-seed sensitivity. The cascade is not evaluated: although its aggregation is deterministic, it would inherit variability from its sampled member outputs. The other four studies use greedy decoding, which is also our recommended setting for reported marks. \textbf{Study 4 (error decomposition)} examines absolute error by subject, answer content, and criterion weight. Judgments are categorized as exact, adjacent over-score, large over-score, adjacent under-score, large under-score, or parse failure. These groups describe assessment items, not protected student attributes. The study therefore evaluates error concentration rather than demographic fairness.


\textbf{Study 5 (explainability)} evaluates a score-conditioned attention map over image patches. Attention at the score-generation position is averaged across the last eight decoder layers, then adjusted by subtracting nearby pre-score attention to reduce shared positional patterns~\cite{abnar2020attention,chefer2021transformer}. The layer range was selected through preliminary inspection; neither that choice nor the subtraction was ablated. Conclusions therefore apply to this configuration. For deletion testing, the highest-attention patches are masked at fractions $\rho\in\{0.1,0.2,0.3,0.5\}$ and the answer is regraded. The absolute score change is
\begin{equation}
\Delta_{\rho} = |\hat{s}_{\text{full}} - \hat{s}_{\text{mask}(\rho)}|,
\label{eq:comprehensiveness}
\end{equation}
and is compared with masking the same number of randomly selected patches. We call the proportion of items for which attention-guided deletion causes a strictly larger score change than the corresponding random control the \textit{faithfulness rate}. This operational measure tests relative sensitivity, not a guarantee of causal explanation. Four faculty reviewers, distinct from the Study 2 markers, also appraised the same ten explained items on five statements: clarity, correspondence with their reading, trust, workflow usefulness, and willingness to adopt. This produced 200 ratings on 50 item--statement units.


The browser-based instrument used in Studies 2 and 5 constrains human marks to $[0,m_j]$ on the same 0.05-mark grid as model scores. It stores progress locally, assigns participant identifiers automatically, and exports one spreadsheet per participant per study, avoiding manual transcription.

\section{Experimental Results}
\label{sec:results}

This section reports findings from five studies on grading accuracy, agreement with human markers, repeated-run reliability, error patterns, and explanation quality.


\subsection{Implementation and Reproducibility}

Experiments used PyTorch~\cite{paszke2019pytorch}, Hugging Face Transformers, and PEFT~\cite{wolf2020transformers,mangrulkar2022peft} with bfloat16 precision and fused AdamW on a single data-center accelerator. Device details, memory use, and run times are recorded in the released manifests. Training used weight decay $10^{-2}$, a cosine schedule with 3\% warmup, gradient clipping at 1.0, three epochs, and an effective batch size of 16. The epoch budget was fixed before evaluation. LoRA used learning rate $10^{-4}$, rank $r=16$, scaling $\alpha=32$, and dropout 0.05. Partial fine-tuning used learning rate $2\times10^{-5}$ and the decoder blocks in Table~\ref{tab:hyper}. Donut used a separate encoder-decoder recipe with a 512-token decoder prompt and full updates in its fourth configuration. Checkpoints were selected by the lowest validation loss, using the 208-record validation split. Data ordering, exemplar selection, and initialization used seed 42. Training was not repeated across seeds, so small differences between configurations should not be treated as established rankings. Inference used greedy decoding with a 256-token generation cap, except for the stochastic repetitions in Study 3. Reproduction also depends on matching the recorded software, hardware, and numerical settings.


\begin{table}[!t]
\vspace{-2pt}
\centering
\caption{Backbones and training configuration. P is parameters in billions and Px the long-side resolution at inference. Batch entries are per-device batch / gradient accumulation, giving an effective batch of 16. Blocks are the decoder blocks left trainable.}
\label{tab:hyper}
\vspace{-5pt}
\setlength{\tabcolsep}{3pt}
\begin{tabular}{lrrccc}
\hline
\textbf{Backbone} & \textbf{P (B)} & \textbf{Px} & \textbf{PFT bs} & \textbf{LoRA bs} & \textbf{Blocks} \\ \hline
Qwen2.5-VL-7B-Instruct & 7.6 & 1280 & 2 / 8 & 4 / 4 & 20--27 \\
InternVL3-8B & 8.0 & 1280 & 2 / 8 & 4 / 4 & 20--27 \\
Pixtral-12B & 12.7 & 768 & 1 / 16 & 2 / 8 & 31--39 \\
Donut-base & 0.2 & 1280 & 4 / 4 & 4 / 4 & all (full FT) \\
\hline
\end{tabular}
\end{table}

\begin{table*}[!htb]
\vspace{-2pt}
\centering
\caption{Agreement panel on the held-out test split (291 criterion-level judgments over 72 answers). QWK, LWK and $\kappa$ are quadratic, linear and unweighted kappa on the 0.05-mark grid; $r_{P}$ and $r_{S}$ are Pearson and Spearman correlation; EM and AM are exact and within-one-mark agreement; PR is the fraction of outputs that parsed. Best value per column in bold.}
\label{tab:metrics-all}
\vspace{-5pt}
\setlength{\tabcolsep}{4pt}
\scriptsize
\begin{tabular}{llrrrrrrrrrrrr}
\hline
\textbf{Setup} & \textbf{Model} & \textbf{QWK} & \textbf{LWK} & \textbf{$\kappa$} & \textbf{$r_{P}$} & \textbf{$r_{S}$} & \textbf{MAE} & \textbf{RMSE} & \textbf{Bias} & \textbf{nMAE} & \textbf{EM} & \textbf{AM} & \textbf{PR} \\ \hline
Zero-shot & Qwen2.5-VL & 0.599 & 0.436 & 0.161 & 0.627 & 0.644 & 0.664 & 0.938 & 0.205 & 0.233 & 0.285 & 0.887 & \textbf{1.00} \\
 & InternVL3 & 0.507 & 0.351 & 0.112 & 0.527 & 0.520 & 0.742 & 1.014 & 0.045 & 0.261 & 0.261 & 0.849 & \textbf{1.00} \\
 & Pixtral & 0.483 & 0.388 & 0.203 & 0.506 & 0.538 & 0.718 & 1.058 & 0.188 & 0.247 & 0.320 & 0.845 & \textbf{1.00} \\
 & Donut & -- & -- & -- & -- & -- & -- & -- & -- & -- & -- & -- & 0.00 \\
 & Cascade & 0.575 & 0.398 & 0.081 & 0.606 & 0.617 & 0.691 & 0.935 & 0.130 & 0.244 & 0.186 & 0.835 & \textbf{1.00} \\
\hline
Few-shot & Qwen2.5-VL & 0.586 & 0.433 & 0.178 & 0.595 & 0.614 & 0.680 & 0.967 & -0.055 & 0.246 & 0.313 & 0.883 & \textbf{1.00} \\
 & InternVL3 & 0.463 & 0.286 & 0.037 & 0.521 & 0.540 & 0.979 & 1.254 & -0.570 & 0.351 & 0.192 & 0.725 & \textbf{1.00} \\
 & Pixtral & 0.432 & 0.315 & 0.136 & 0.448 & 0.467 & 0.796 & 1.091 & -0.052 & 0.280 & 0.258 & 0.784 & \textbf{1.00} \\
 & Donut & -- & -- & -- & -- & -- & -- & -- & -- & -- & -- & -- & 0.00 \\
 & Cascade & 0.570 & 0.365 & 0.056 & 0.601 & 0.627 & 0.787 & 1.007 & -0.303 & 0.286 & 0.151 & 0.742 & \textbf{1.00} \\
\hline
LoRA & Qwen2.5-VL & 0.727 & \textbf{0.663} & \textbf{0.537} & 0.727 & 0.729 & \textbf{0.435} & 0.847 & 0.014 & \textbf{0.150} & \textbf{0.608} & \textbf{0.938} & \textbf{1.00} \\
 & InternVL3 & 0.707 & 0.649 & 0.527 & 0.708 & 0.733 & 0.448 & 0.869 & \textbf{-0.003} & 0.155 & 0.601 & \textbf{0.938} & \textbf{1.00} \\
 & Pixtral & 0.659 & 0.602 & 0.473 & 0.660 & 0.681 & 0.511 & 0.943 & -0.080 & 0.169 & 0.557 & 0.931 & \textbf{1.00} \\
 & Donut$^{\dagger}$ & 0.080 & 0.057 & 0.016 & 0.108 & 0.125 & 0.933 & 1.221 & -0.018 & 0.313 & 0.237 & 0.724 & 0.52 \\
 & Cascade$^{\ddagger}$ & \textbf{0.732} & 0.651 & 0.455 & \textbf{0.733} & \textbf{0.743} & 0.442 & \textbf{0.820} & 0.003 & 0.152 & 0.526 & 0.931 & \textbf{1.00} \\
\hline
Partial FT & Qwen2.5-VL & 0.704 & 0.622 & 0.470 & 0.710 & 0.714 & 0.487 & 0.883 & 0.148 & 0.158 & 0.546 & \textbf{0.938} & \textbf{1.00} \\
/ full & InternVL3 & 0.629 & 0.608 & 0.506 & 0.632 & 0.646 & 0.510 & 0.996 & 0.102 & 0.164 & 0.581 & 0.931 & \textbf{1.00} \\
 & Pixtral & 0.552 & 0.513 & 0.407 & 0.562 & 0.595 & 0.657 & 1.129 & -0.205 & 0.227 & 0.495 & 0.859 & \textbf{1.00} \\
 & Donut$^{\dagger}$ & 0.295 & 0.222 & 0.119 & 0.372 & 0.393 & 0.763 & 1.070 & 0.090 & 0.272 & 0.327 & 0.820 & 0.52 \\
 & Cascade$^{\ddagger}$ & 0.712 & 0.624 & 0.398 & 0.717 & 0.722 & 0.477 & 0.851 & 0.116 & 0.154 & 0.467 & 0.931 & \textbf{1.00} \\
\hline
\multicolumn{14}{l}{\footnotesize $^{\dagger}$Conditional on the parsed subset. $^{\ddagger}$Quantized to a 0.25-mark grid. Donut's fourth cell is full fine-tuning.} \\
\hline
\end{tabular}
\end{table*}


\subsection{Study 1: Accuracy}

\begin{figure}[tp]
\centering
\includegraphics[width=\columnwidth]{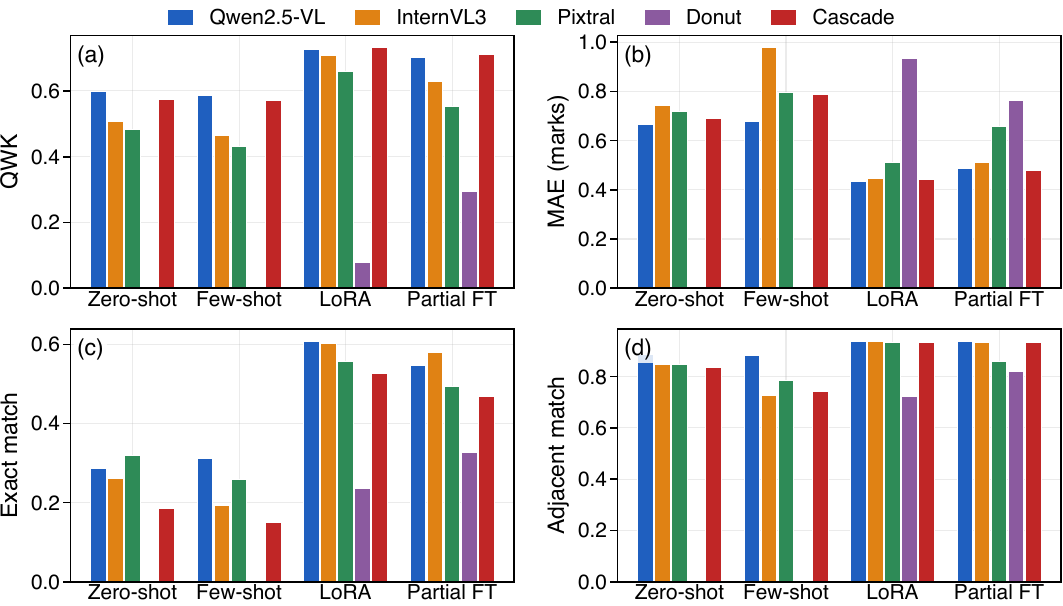}
\vspace{-15pt}
\caption{Study 1 accuracy across the 20 systems: (a) quadratic weighted kappa, (b) mean absolute error in marks, (c) exact match, and (d) adjacent match.}
\label{fig:study1}
\end{figure}

Study 1 evaluates agreement with the examiner on 291 held-out criteria. Table~\ref{tab:metrics-all} presents the full metric panel, Table~\ref{tab:deltas} compares adaptation with zero-shot prompting, and Fig.~\ref{fig:study1} summarizes accuracy. The main finding is that adaptation consistently improves the three instruction-tuned VLMs, while model selection depends on the scoring properties required by the assessment task.


Qwen2.5-VL with LoRA is the strongest single backbone by QWK, achieving 0.727 with MAE 0.435 marks, exact match 0.608, and adjacent match 0.938. The LoRA cascade reaches QWK 0.732, but its MAE is higher (0.442) and exact match lower (0.526). Qwen2.5-VL also leads the cascade on linear and unweighted kappa, nMAE, and adjacent match; the cascade leads on correlations, RMSE, and absolute bias. We select Qwen2.5-VL with LoRA as the reference system because lower absolute error and higher exact agreement are directly relevant to criterion-level marking. The small QWK difference does not establish a reliable ordering, particularly given the cascade's coarser output grid. LoRA improves QWK by 0.128--0.200 across the three VLMs, compared with 0.069--0.122 for partial fine-tuning. Cascade gains are 0.157 and 0.137, respectively. LoRA exceeds partial fine-tuning for every VLM and the cascade; Donut performs better with full fine-tuning. Model size alone does not rank the results: Pixtral-12B with LoRA reaches 0.659, compared with 0.727 for Qwen2.5-VL-7B. This comparison reflects the tested resource settings, including Pixtral's lower image resolution.


Few-shot prompting reduces QWK by 0.005--0.051 in all four configurations with valid prompted scores. Exact match falls for InternVL3 (0.261 to 0.192), Pixtral (0.320 to 0.258), and the cascade (0.186 to 0.151), but rises for Qwen2.5-VL (0.285 to 0.313). Worked examples do not improve QWK under the tested prompt and context budgets, although their effects differ by metric. Donut produces no parsable marks under either prompted label and approximately 52\% coverage after adaptation. Its QWK values of 0.080 with LoRA and 0.295 with full fine-tuning are conditional on the parsed subsets. Coverage remains an essential part of model comparison: agreement on parsed outputs cannot represent a grader's performance on the complete assessment set.


\begin{table}[!t]
\centering
\caption{Change relative to the zero-shot setting of the same backbone. Positive $\Delta$QWK and negative $\Delta$MAE are improvements.}
\label{tab:deltas}
\vspace{-6pt}
\setlength{\tabcolsep}{4pt}
\scriptsize
\begin{tabular}{llrrrr}
\hline
\textbf{Model} & \textbf{Setup} & \textbf{QWK} & \textbf{$\Delta$QWK} & \textbf{MAE} & \textbf{$\Delta$MAE} \\ \hline
Qwen2.5-VL & Few-shot & 0.586 & -0.013 & 0.680 & +0.016 \\
Qwen2.5-VL & LoRA & 0.727 & +0.128 & 0.435 & -0.229 \\
Qwen2.5-VL & Partial FT & 0.704 & +0.104 & 0.487 & -0.177 \\
\hline
InternVL3 & Few-shot & 0.463 & -0.044 & 0.979 & +0.236 \\
InternVL3 & LoRA & 0.707 & +0.200 & 0.448 & -0.295 \\
InternVL3 & Partial FT & 0.629 & +0.122 & 0.510 & -0.233 \\
\hline
Pixtral & Few-shot & 0.432 & -0.051 & 0.796 & +0.078 \\
Pixtral & LoRA & 0.659 & +0.176 & 0.511 & -0.206 \\
Pixtral & Partial FT & 0.552 & +0.069 & 0.657 & -0.061 \\
\hline
Donut & Few-shot & -- & -- & -- & -- \\
Donut & LoRA & 0.080 & n/a & 0.933 & n/a \\
Donut & Full FT & 0.295 & n/a & 0.763 & n/a \\
\hline
Cascade & Few-shot & 0.570 & -0.005 & 0.787 & +0.096 \\
Cascade & LoRA & 0.732 & +0.157 & 0.442 & -0.249 \\
Cascade & Partial FT & 0.712 & +0.137 & 0.477 & -0.214 \\
\hline
\end{tabular}
\end{table}


\begin{table*}[!htb]
\vspace{-2pt}
\centering
\caption{Quadratic weighted kappa per question template, grouped by subject and annotated with the dominant answer content. $n$ is the number of criterion-level judgments the question contributes to the test split; values are pooled over the criteria of the question.}
\label{tab:subject-qwk}
\vspace{-5pt}
\setlength{\tabcolsep}{2.2pt}
\scriptsize
\resizebox{\textwidth}{!}{%
\begin{tabular}{llrrrrrrrrrrrrrrrrrrr}
\hline
 &  &  & \multicolumn{4}{c}{\textbf{Zero-shot}} & \multicolumn{4}{c}{\textbf{Few-shot}} & \multicolumn{5}{c}{\textbf{LoRA}} & \multicolumn{5}{c}{\textbf{Partial FT}} \\
\textbf{Subject} & \textbf{Q} & \textbf{n} & Qwen & Intern & Pixt & Casc & Qwen & Intern & Pixt & Casc & Qwen & Intern & Pixt & Don & Casc & Qwen & Intern & Pixt & Don & Casc \\ \hline
Machine Learning (Text) & Q1 & 8 & .00 & -0.06 & -0.06 & -0.05 & .27 & .14 & -0.08 & .24 & .45 & .70 & .84 & .00 & .61 & .49 & .81 & .86 & .00 & .73 \\
 & Q2 & 36 & .07 & .41 & .64 & .25 & .00 & .58 & .47 & .43 & .21 & .69 & .63 & -0.04 & .61 & .12 & .71 & .62 & .00 & .60 \\
Digital Image Processing (Text) & Q3 & 12 & .32 & .11 & .05 & .24 & .32 & .26 & .21 & .29 & .00 & .00 & .53 & .40 & .00 & .00 & .00 & .00 & .00 & .00 \\
 & Q4 & 16 & .13 & .00 & .00 & .06 & .20 & .07 & .00 & .13 & .52 & .60 & .37 & .03 & .58 & .00 & .62 & -0.05 & .06 & .45 \\
Database Systems (Text+Tab+Eq) & Q5 & 40 & .48 & .27 & .51 & .40 & .41 & .00 & .03 & .14 & .85 & .83 & .53 & -0.16 & .87 & .76 & .68 & .20 & -0.05 & .77 \\
 & Q6 & 24 & .58 & .53 & .57 & .61 & .28 & .33 & .44 & .38 & .60 & .51 & .60 & .09 & .60 & .51 & .50 & .51 & .00 & .55 \\
Computer Networks (Text+Diag) & Q7 & 45 & .81 & .77 & .79 & .82 & .82 & .63 & .66 & .79 & .78 & .79 & .79 & .36 & .81 & .81 & .81 & .68 & .49 & .84 \\
Data Mining (Text) & Q8 & 32 & .56 & .62 & .16 & .60 & .74 & .59 & .21 & .74 & .89 & .85 & .77 & -0.03 & .88 & .88 & .76 & .49 & .00 & .87 \\
Algorithms (Text+Eq+Diag) & Q9 & 40 & -0.12 & .03 & -0.24 & -0.06 & -0.07 & .07 & .09 & .04 & .30 & .30 & .24 & .01 & .33 & .27 & .46 & .00 & .11 & .39 \\
Fisheries (Text) & Q10 & 18 & .05 & -0.03 & .26 & -0.01 & -0.12 & -0.04 & .06 & -0.06 & .27 & .25 & .25 & .32 & .27 & .07 & .07 & .13 & -- & .07 \\
Object Oriented Prog. (Text+Code) & Q11 & 20 & .03 & .08 & .04 & .06 & -0.01 & .17 & .12 & .07 & .09 & -0.07 & -0.11 & .00 & -0.08 & .12 & -0.35 & .11 & -0.05 & -0.11 \\
\hline
\textbf{Total} & & \textbf{291} & & & & & & & & & & & & & & & & & & \\
\hline
\end{tabular}}
\end{table*}


Template-level results distinguish agreement from absolute error (Table~\ref{tab:subject-qwk}). For Qwen2.5-VL with LoRA, QWK is highest on Data Mining Q8 (0.89), Database Systems Q5 (0.85), and Computer Networks Q7 (0.78), and lower on Digital Image Processing Q3 (0.00), Object Oriented Programming Q11 (0.09), Machine Learning Q2 (0.21), and Fisheries Q10 (0.27). Fisheries combines low QWK with MAE of approximately 0.12 marks (Table~\ref{tab:subject-mae}), illustrating the need to interpret agreement alongside error magnitude. Score distributions can affect kappa, although subject means do not establish the variance within each test template. Q11 combines low agreement with the largest subject-level error for VLM configurations. These estimates, based on 8--45 judgments per template, identify cases for closer assessment review.


\subsection{Study 2: Comparison with Human Agreement}

\begin{figure*}[tp]
\centering
\includegraphics[width=0.7\textwidth]{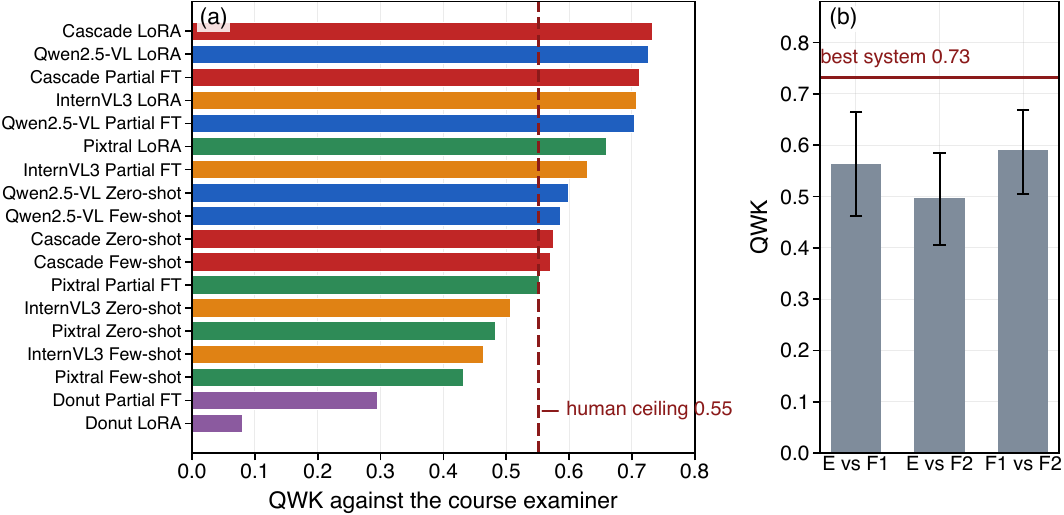}
\vspace{-10pt}
\caption{Study 2. (a) Agreement of every system with the course examiner against the measured human agreement baseline. (b) Pairwise agreement among the three human sources with bootstrap 95\% intervals; E is the course examiner, R1 and R2 the independent faculty graders.}
\label{fig:study2}
\end{figure*}

Study 2 compares model--examiner agreement with agreement among qualified human markers on the same 291 criteria. Two additional faculty markers independently regraded the complete test split using the browser booklet described in Section~\ref{sec:method}. Table~\ref{tab:human} and Fig.~\ref{fig:study2} report the comparisons.


Human-pair QWK values are 0.564, 0.497, and 0.592, with mean $\bar{\kappa}_H=0.551$ and pairwise MAE of 0.687--0.746 marks. The independent markers agree exactly on 19.6\% of criteria. Five model point estimates exceed the largest human-pair bootstrap upper bound of 0.67: the LoRA cascade (0.732), Qwen2.5-VL with LoRA (0.727), the partial-fine-tuning cascade (0.712), InternVL3 with LoRA (0.707), and Qwen2.5-VL with partial fine-tuning (0.704). The reference system's MAE of 0.435 is 0.252 marks below the lowest human-pair MAE. These comparisons show that adapted models can closely reproduce the examiner's scoring pattern. They also identify the role of calibration: models learned from 1,483 examiner-assigned training marks, whereas the independent markers applied the rubric without that training. The bootstrap-bound comparison is descriptive and concerns agreement with a particular marking standard.


\begin{table}[!t]
\vspace{-2pt}
\centering
\caption{Pairwise agreement among the three human markers and, for reference, the three strongest automated systems against the course examiner. E is the course examiner; R1 and R2 are the independent faculty markers. Brackets give bootstrap 95\% intervals for QWK.}
\label{tab:human}
\vspace{-5pt}
\setlength{\tabcolsep}{3pt}
\begin{tabular}{llrrr}
\hline
\textbf{Pair} & \textbf{QWK [95\% CI]} & \textbf{MAE} & \textbf{EM} & \textbf{AM} \\ \hline
E vs R1 & 0.564 [0.46, 0.66] & 0.687 & 0.330 & 0.849 \\
E vs R2 & 0.497 [0.40, 0.58] & 0.746 & 0.237 & 0.835 \\
R1 vs R2 & 0.592 [0.51, 0.67] & 0.688 & 0.196 & 0.856 \\
\hline
\textit{Mean pairwise} $\bar{\kappa}_{H}$ & \textit{0.551} & & & \\
\textit{Highest upper bound} & \textit{0.67} & & & \\ \hline
Qwen2.5-VL LoRA vs E & 0.727 & 0.435 & 0.608 & 0.938 \\
Cascade LoRA vs E & 0.732 & 0.442 & 0.526 & 0.931 \\
Cascade Partial FT vs E & 0.712 & 0.477 & 0.467 & 0.931 \\
\hline
\end{tabular}
\end{table}


\subsection{Study 3: Run-to-Run Reliability}

\begin{figure*}[tp]
\centering
\includegraphics[width=0.6\textwidth]{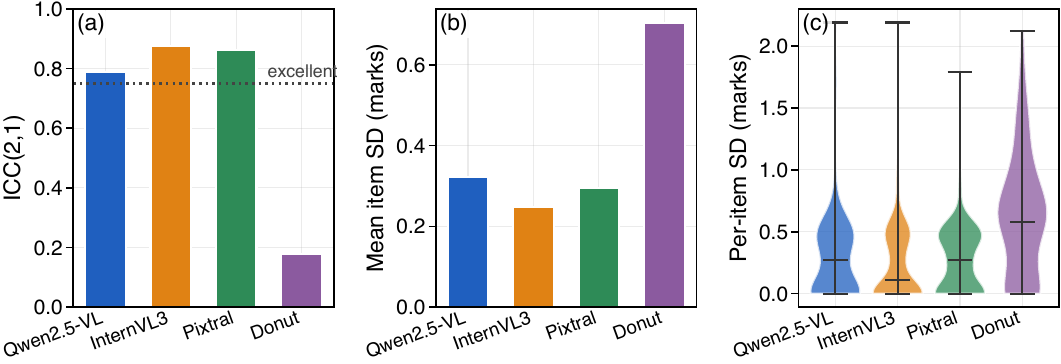}
\vspace{-10pt}
\caption{Study 3 reliability over five stochastic repetitions: (a) ICC(2,1) with a 0.75 reference line, (b) mean per-item standard deviation, and (c) the distribution of per-item standard deviations.}
\label{fig:study3}
\end{figure*}

Study 3 measures variation over five stochastic repetitions per LoRA-adapted backbone. Reliability complements agreement when evaluating a scoring system~\cite{landis1977measurement,williamson2012framework}. ICC(2,1) is 0.874 for InternVL3, 0.862 for Pixtral, and 0.790 for Qwen2.5-VL, each over all 291 test criteria. Their mean per-item standard deviations are 0.249, 0.295, and 0.322 marks, respectively. These ICC values exceed the 0.75 reference line plotted in Fig.~\ref{fig:study3}, but that line is not a criterion for deployment readiness. Exact repeatability is only 49.8\% for InternVL3, 36.4\% for Pixtral, and 41.9\% for Qwen2.5-VL. Donut has ICC 0.177 and mean per-item standard deviation 0.703 marks on just nine criteria parsed in every run; this selected sample does not support a comparable reliability estimate. Table~\ref{tab:reliability} summarizes the results. The most accurate single backbone is not the most stable under sampling. Greedy decoding avoids this source of variation, although training and numerical reproducibility require separate checks.


\begin{table}[!t]
\vspace{-2pt}
\centering
\caption{Run-to-run reliability of the four backbones under LoRA adaptation over five stochastic repetitions at temperature 0.7, varying the decoding seed only. $\bar{s}$ is the mean per-item standard deviation in marks and ID the share of items scored identically in all five runs. Donut is computed on the 9 criteria it parsed in all five runs.}
\label{tab:reliability}
\vspace{-5pt}
\setlength{\tabcolsep}{3pt}
\begin{tabular}{lrrrrr}
\hline
\textbf{Model} & \textbf{Items} & \textbf{ICC(2,1)} & \textbf{$\bar{s}$} & \textbf{Range} & \textbf{ID} \\ \hline
Qwen2.5-VL & 291 & 0.790 & 0.322 & 0.683 & 0.419 \\
InternVL3 & 291 & 0.874 & 0.249 & 0.525 & 0.498 \\
Pixtral & 291 & 0.862 & 0.295 & 0.630 & 0.364 \\
Donut & 9 & 0.177 & 0.703 & 1.556 & 0.000 \\
\hline
\end{tabular}
\end{table}


\subsection{Study 4: Error Concentration}

\begin{figure*}[tp]
\centering
\includegraphics[width=0.7\textwidth]{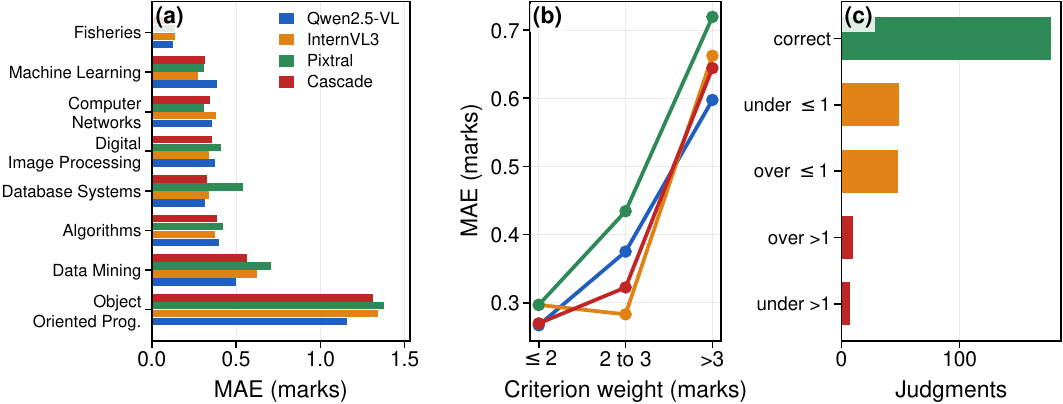}
\vspace{-10pt}
\caption{Study 4 error decomposition under LoRA adaptation: (a) MAE by subject, (b) MAE by criterion mark weight, and (c) failure taxonomy of the strongest system. Eight subjects appear rather than the nine of Table~\ref{tab:dataset}, since the E-commerce template contributes no test judgment.}
\label{fig:study4}
\end{figure*}

Study 4 examines error concentration by subject, criterion weight, and failure category (Fig.~\ref{fig:study4} and Table~\ref{tab:subject-mae}). For Qwen2.5-VL with LoRA, subject MAE ranges from 0.125 marks in Fisheries to 1.625 in Object Oriented Programming, a gap of 1.500 marks. The corresponding gaps are 1.712 for the cascade and 1.736 for InternVL3 with LoRA. Object Oriented Programming has the highest subject MAE in sixteen of the eighteen configurations with valid scores. The exceptions are Donut with LoRA and full fine-tuning, whose largest errors occur in Algorithms and Fisheries, respectively. This result concerns one code-bearing template with 20 test judgments and should not be generalized to programming assessment.


For the reference system, MAE rises from 0.267 marks on 92 criteria worth at most 2 marks, to 0.375 on 76 criteria worth more than 2 and at most 3, and 0.598 on 123 criteria worth more than 3. Larger maxima permit larger absolute errors. The overall nMAE of 0.150 does not establish whether proportional error also rises across these groups; that requires group-specific normalized errors. Heavily weighted criteria nevertheless deserve attention because their errors can have larger effects on aggregated attainment. Of all judgments, 60.8\% match the examiner exactly, 33.0\% differ by up to one mark, and 6.2\% differ by more than one mark. Overall signed bias is only $+0.014$ marks, but this near-zero average can conceal substantial errors for individual students. Individual-mark review is therefore more informative than checking cohort totals alone.


\begin{table*}[!htb]
\centering
\caption{Mean absolute error in marks per subject for every system.}
\label{tab:subject-mae}
\vspace{-6pt}
\setlength{\tabcolsep}{2.2pt}
\scriptsize
\resizebox{\textwidth}{!}{%
\begin{tabular}{lrrrrrrrrrrrrrrrrrr}
\hline
\textbf{Subject} & Qwen2 & Inter & Pixtr & Casca & Qwen2 & Inter & Pixtr & Casca & Qwen2 & Inter & Pixtr & Donut & Casca & Qwen2 & Inter & Pixtr & Donut & Casca \\
 & Zero & Zero & Zero & Zero & Few- & Few- & Few- & Few- & LoRA & LoRA & LoRA & LoRA & LoRA & Part & Part & Part & Part & Part \\ \hline
Algorithms & .57 & .88 & .55 & .69 & .57 & 1.12 & .42 & .76 & .40 & .38 & .42 & 1.79 & .39 & .47 & .33 & .42 & 1.08 & .40 \\
Computer Networks & .48 & .47 & .45 & .46 & .44 & .66 & .59 & .50 & .36 & .38 & .31 & .74 & .34 & .31 & .33 & .53 & .49 & .31 \\
Data Mining & .94 & .84 & 1.38 & .90 & .70 & .88 & 1.28 & .67 & .38 & .47 & .53 & 1.50 & .42 & .41 & .59 & 1.00 & 1.43 & .45 \\
Database Systems & .51 & .68 & .60 & .57 & .69 & 1.35 & .82 & 1.01 & .32 & .34 & .54 & .76 & .33 & .37 & .41 & .81 & .46 & .38 \\
Digital Image Processing & .71 & .84 & .86 & .78 & .77 & .96 & 1.00 & .87 & .38 & .34 & .41 & .50 & .36 & .48 & .38 & .59 & .55 & .43 \\
Fisheries & .51 & .54 & .32 & .51 & .58 & 1.06 & .43 & .78 & .12 & .14 & .14 & .44 & .12 & .14 & .14 & .21 & 2.00 & .14 \\
Machine Learning & .50 & .52 & .45 & .51 & .47 & .40 & .59 & .41 & .39 & .27 & .31 & .57 & .32 & .40 & .26 & .26 & .52 & .31 \\
Object Oriented Prog. & 1.73 & 1.68 & 1.73 & 1.70 & 1.77 & 1.62 & 1.62 & 1.68 & 1.62 & 1.88 & 1.93 & 1.78 & 1.84 & 1.93 & 2.52 & 1.73 & 1.79 & 2.12 \\
\hline
\end{tabular}}
\end{table*}


\subsection{Study 5: Explainability}

Template-level results distinguish agreement from absolute error (Table~\ref{tab:subject-qwk}). For Qwen2.5-VL with LoRA, QWK is highest on Data Mining Q8 (0.89), Database Systems Q5 (0.85), and Computer Networks Q7 (0.78), and lower on Digital Image Processing Q3 (0.00), Object Oriented Programming Q11 (0.09), Machine Learning Q2 (0.21), and Fisheries Q10 (0.27). Fisheries combines low QWK with MAE of approximately 0.12 marks (Table~\ref{tab:subject-mae}), illustrating the need to interpret agreement alongside error magnitude. Score distributions can affect kappa, although subject means do not establish the variance within each template. Q11 combines low agreement with the largest subject-level error for VLM configurations. These estimates, based on 8--45 judgments per template, identify cases for assessment review.


The four faculty reviewers differ substantially in their appraisals (Table~\ref{tab:survey}). Median ratings range from 3.0 to 3.5 on the five-point scale. Favorable ratings never exceed 50\%, and willingness to adopt is lowest at 33\%. ICC(2,1) across the 50 item--statement units is approximately $-0.03$, with reviewer means ranging from about 1.7 to 4.3. These values indicate little agreement among reviewers. Given the four-person panel and repeated ratings of ten items, the results describe expert appraisal rather than population acceptance.


\begin{table}[!t]
\vspace{-2pt}
\centering
\caption{Expert appraisal of explanations by four faculty reviewers (F1--F4), a panel distinct from markers R1--R2 of Table~\ref{tab:human}. Each rated the ten explained items on five statements. Md is the median and Fav.\ the share of ratings 4 or 5.}
\label{tab:survey}
\vspace{-5pt}
\setlength{\tabcolsep}{3pt}
\begin{tabular}{lrrrrrr}
\hline
\textbf{Statement} & \textbf{Md} & \textbf{Fav.} & \textbf{F1} & \textbf{F2} & \textbf{F3} & \textbf{F4} \\ \hline
Explanation is clear & 3.5 & 0.50 & 1.8 & 3.4 & 4.3 & 3.5 \\
Regions match my own reading & 3.5 & 0.50 & 1.7 & 3.2 & 4.0 & 3.9 \\
Trust in the awarded mark & 3.0 & 0.42 & 1.4 & 3.0 & 4.2 & 3.5 \\
Useful in my grading workflow & 3.0 & 0.47 & 1.7 & 3.1 & 4.6 & 3.7 \\
Would adopt with teacher override & 3.0 & 0.33 & 1.7 & 2.9 & 4.6 & 3.3 \\
\hline
\multicolumn{7}{l}{\footnotesize ICC(2,1) across the four reviewers over $n=50$ units $=-0.027$.} \\ \hline
\end{tabular}
\end{table}

\begin{table}[!t]
\vspace{-2pt}
\centering
\caption{Deletion faithfulness of the score-conditioned attention maps on 30 test criteria. Compr.\ is the mark change after masking the highest-attention patches and Random masks the same number uniformly; Faithful is the share of items whose attention-guided change exceeds their own random control; $p$ is the paired Wilcoxon value between the two columns.}
\label{tab:faithfulness}
\vspace{-5pt}
\setlength{\tabcolsep}{3pt}
\begin{tabular}{rrrrrr}
\hline
\textbf{Mask} & \textbf{n} & \textbf{Compr.} & \textbf{Random} & \textbf{Faithful} & \textbf{$p$} \\ \hline
10\% & 30 & 0.367 & 0.283 & 0.100 & 0.102 \\
20\% & 30 & 0.500 & 0.417 & 0.233 & 0.119 \\
30\% & 30 & 0.683 & 0.517 & 0.167 & 0.206 \\
50\% & 30 & 0.617 & 0.817 & 0.100 & 0.248 \\
\hline
\end{tabular}
\end{table}


\begin{figure*}[tp]
\centering
\includegraphics[width=0.64\textwidth]{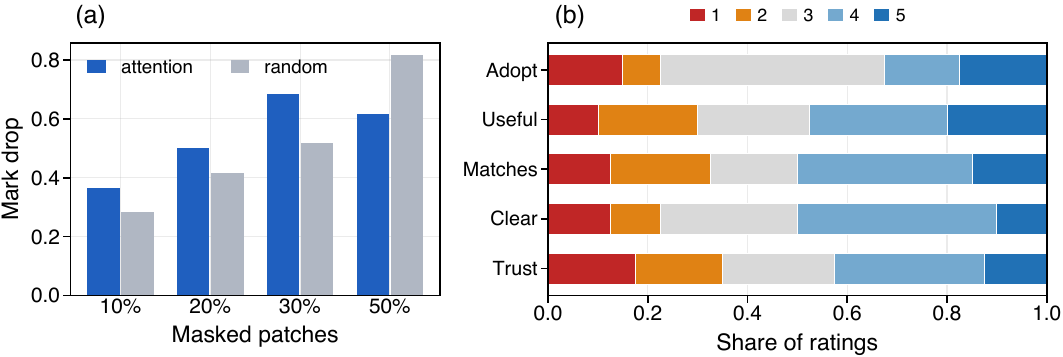}
\vspace{-10pt}
\caption{Study 5. (a) Deletion faithfulness of score-conditioned attention against a random-masking control at four masked fractions, on 30 test criteria. (b) Distribution of the four reviewers' ratings across the five appraisal statements.}
\label{fig:study5}
\end{figure*}


\begin{figure}[!t]
\centering
\includegraphics[width=0.8\columnwidth]{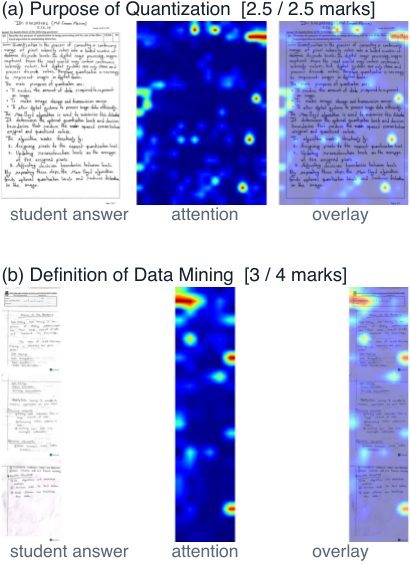}
\vspace{-10pt}
\caption{Score-conditioned attention on two correctly graded criteria, with the awarded mark and the criterion maximum in brackets. Each row shows the student answer, the attention map, and their overlay.}
\label{fig:heatmaps}
\end{figure}


\subsection{Implications for Assessment-Tool Design}

The five studies identify requirements for criterion-level assessment: calibration to a rubric, consistent production of individual marks, targeted review, and evidence that explanations reflect scoring behavior. Their value is that success on one requirement does not establish success on the others.

\textbf{Calibration and model selection (RQ1--RQ2):} LoRA improves agreement across the three instruction-tuned VLMs, supporting adaptation to the local scoring task. Model selection should consider the criterion-level decision: Qwen2.5-VL has lower MAE and higher exact agreement than the cascade despite the cascade's slightly higher QWK. Human agreement also depends on rubric specificity and rater training~\cite{jonsson2007rubrics}. The human comparison therefore provides a reference for interpreting examiner alignment, with the models' access to training marks made explicit. For OBE tools, retaining each criterion's identity and maximum keeps predicted marks compatible with outcome mappings and enables educators to inspect individual scoring decisions.

\textbf{Repeatability of reported marks (RQ3):} ICC values of 0.790--0.874 coexist with changes on 50.2--63.6\% of criteria across sampled runs. Aggregate reliability can conceal variation in a mark returned to an individual student. Greedy decoding avoids this sampling variability and provides a default for reported scores. Exact repeatability should accompany aggregate reliability when assessing a grading tool.

\textbf{Review guided by error concentration (RQ4):} The reference system's subject MAE gap is 1.500 marks, with its largest error on the code-bearing template. Absolute error also rises with criterion weight. These patterns support prioritizing such cases for teacher review within this assessment setting. They also show why a small cohort-level bias is insufficient for judging the quality of individual marks. Retaining criterion-level outputs enables review before those marks contribute to CLO attainment.

\textbf{Explanations evaluated separately from scores (RQ5):} The attention maps show no significant advantage over random masking, and faculty appraisals lack consensus. A plausible visualization should not be treated as validation of the mark. For tool design, the scoring decision and its explanation require separate evidence: agreement establishes alignment with a marking source, whereas faithfulness and expert appraisal address whether the explanation can support review.

\subsection{Scope and Limitations}

The findings concern one institution, twelve question templates, and a test split covering eight subjects and eleven templates. They establish performance on held-out answers within this corpus; transfer to unseen templates or institutions remains outside the evaluation. Training labels come from one examiner, and the human baseline includes two additional markers. Human confidence intervals use criterion-level resampling and do not account for dependence among criteria from the same answer. Training uses one seed, while Study 3 varies decoding seeds only. Explanation evidence is limited to one attribution configuration, 30 deletion-test items, and four reviewers. The resulting design implications guide the development of assessment tools; the study does not measure changes in teacher workload or student learning.

\section{Conclusion}
\label{sec:conclusion}


This study provides an integrated evaluation of VLMs for criterion-level grading of handwritten examinations in OBE. Across 20 configurations and 291 held-out judgments, Qwen2.5-VL with LoRA achieves QWK 0.727 and MAE 0.435 marks against the examiner. Independent human marking places this result in the context of agreement among qualified markers and highlights the role of calibration to the examiner's scoring standard. The combined studies reveal three consequential distinctions: the model with the highest QWK need not have the lowest criterion-level error; strong aggregate reliability can coexist with changing individual marks; and accurate scores can accompany explanations whose faithfulness is not demonstrated.


These findings provide guidance for assessment-tool development: adapt models to the rubric and marking context, use repeatable decoding for reported marks, retain criterion-level records for teacher review, and validate explanations independently. The public implementation and study outputs support reuse of this evaluation framework. Future work examines its transfer across assessment settings and evaluates alternative explanations and classroom use. Within the scope, the contribution is evidence for designing and evaluating teacher-supervised grading tools that preserve the criterion-level structure required by outcome-based assessment.

\section*{Disclosure}


\textbf{Funding:} This work was funded by the Bangladesh Accreditation Council (BAC) under the project \emph{AI-Powered Automated Exam Evaluation and OBE-Based University Assessment System}.


\textbf{Acknowledgment:} The authors thank the participating students, the faculty who regraded the test split and appraised the explanations, the institution whose scripts form the corpus, and the OBE expert who supported rubric preparation.


\textbf{Data and code availability:} The corpus is described in a separate publication~\cite{obedataset2026}. The public repository at \url{https://github.com/Khalidsyfullah/Vision-Language-Models-for-Outcome-Based-Assessment} contains the implementation, configuration files, prompt code, human-rating data, released predictions and study outputs, and scripts used to generate the reported figures. The browser-based instrument used for human marking and expert appraisal is available at \url{https://khalidsyfullah.github.io/obe-teacher-study/}.


\textbf{Ethics:} Answer scripts were collected under written permission from the institution that holds them and were de-identified before processing. Faculty participation in Studies 2 and 5 was voluntary and unremunerated, and participants were identified only by an automatically generated code.


\textbf{Generative AI usage:} Generative artificial intelligence tools, including ChatGPT and Claude, were used to assist with manuscript preparation, language editing, code organization, documentation, and refactoring. The authors independently verified the data, analyses, and reported results and take full responsibility for the research design, interpretation of the findings, and final manuscript.

\bibliographystyle{IEEEtran}
\bibliography{References}

@misc{obedataset2026,
  title={Multimodal examination answer data with expert-designed Outcome-Based Education rubrics for criterion-level assessment},
  author={Jahangir Alam, S. M. and Syfullah, Md Khalid and Ahmed, Saad and Mou, Munira Akter and Rahman, A K Z Rasel and Rahman, A. K. M. Masudur and Ali, Mohammed Sowket},
  year={2026},
  eprint={2608.22346},
  archivePrefix={arXiv},
  primaryClass={cs.CV},
  url={https://arxiv.org/abs/2608.22346}
}

@inproceedings{abnar2020attention,
  title={Quantifying attention flow in transformers},
  author={Abnar, Samira and Zuidema, Willem},
  booktitle={Proceedings of the 58th Annual Meeting of the Association for Computational Linguistics},
  pages={4190--4197},
  year={2020}
}

@article{attali2006erater,
  title={Automated essay scoring with e-rater{\textregistered} V. 2},
  author={Attali, Yigal and Burstein, Jill},
  journal={The Journal of Technology, Learning and Assessment},
  volume={4},
  number={3},
  year={2006}
}

@misc{bai2025qwen25vl,
      title={{Qwen2.5-VL} Technical Report}, 
      author={Shuai Bai and Keqin Chen and Xuejing Liu and Jialin Wang and Wenbin Ge and Sibo Song and Kai Dang and Peng Wang and Shijie Wang and Jun Tang and Humen Zhong and Yuanzhi Zhu and Mingkun Yang and Zhaohai Li and Jianqiang Wan and Pengfei Wang and Wei Ding and Zheren Fu and Yiheng Xu and Jiabo Ye and Xi Zhang and Tianbao Xie and Zesen Cheng and Hang Zhang and Zhibo Yang and Haiyang Xu and Junyang Lin},
      year={2025},
      eprint={2502.13923},
      archivePrefix={arXiv},
      primaryClass={cs.CV},
      url={https://arxiv.org/abs/2502.13923}, 
}

@article{biderman2024lora,
  title={{LoRA} learns less and forgets less},
  author={Biderman, Dan and Portes, Jacob and Ortiz, Jose Javier Gonzalez and Paul, Mansheej and Greengard, Philip and Jennings, Connor and King, Daniel and Havens, Sam and Chiley, Vitaliy and Frankle, Jonathan and others},
  journal={arXiv preprint arXiv:2405.09673},
  year={2024}
}

@book{biggs2011teaching,
  title={Teaching for Quality Learning at University: What the Student Does},
  author={Biggs, John and Tang, Catherine},
  edition={4th},
  publisher={Open University Press},
  address={Maidenhead, U.K.},
  year={2011}
}

@inproceedings{chefer2021transformer,
  title={Transformer interpretability beyond attention visualization},
  author={Chefer, Hila and Gur, Shir and Wolf, Lior},
  booktitle={2021 IEEE/CVF Conference on Computer Vision and Pattern Recognition (CVPR)},
  pages={782--791},
  year={2021},
  organization={IEEE}
}

@article{cohen1968weighted,
  title={Weighted kappa: Nominal scale agreement with provision for scaled disagreement or partial credit},
  author={Cohen, Jacob},
  journal={Psychological Bulletin},
  volume={70},
  number={4},
  pages={213--220},
  year={1968}
}

@inproceedings{cozma2018essay,
  title={Automated essay scoring with string kernels and word embeddings},
  author={Cozma, Madalina and Butnaru, Andrei M and Ionescu, Radu Tudor},
  booktitle={Proceedings of the 56th Annual Meeting of the Association for Computational Linguistics},
  volume={2},
  pages={503--509},
  year={2018}
}

@article{crosilla2025htrllm,
  title={Benchmarking large language models for handwritten text recognition},
  author={Crosilla, Giorgia and Klic, Lukas and Colavizza, Giovanni},
  journal={Journal of Documentation},
  volume={81},
  number={7},
  pages={334--354},
  year={2025},
  publisher={Emerald Publishing Limited}
}

@article{dettmers2023qlora,
  title={{QLoRA}: Efficient finetuning of quantized {LLMs}},
  author={Dettmers, Tim and Pagnoni, Artidoro and Holtzman, Ari and Zettlemoyer, Luke},
  journal={Advances in Neural Information Processing Systems},
  volume={36},
  pages={10088--10115},
  year={2023}
}

@inproceedings{dong2016automatic,
  title={Automatic features for essay scoring: An empirical study},
  author={Dong, Fei and Zhang, Yue},
  booktitle={Proceedings of the 2016 Conference on Empirical Methods in Natural Language Processing},
  pages={1072--1077},
  year={2016}
}

@article{harden2002learning,
  title={Learning outcomes and instructional objectives: Is there a difference?},
  author={Harden, Ronald M},
  journal={Medical Teacher},
  volume={24},
  number={2},
  pages={151--155},
  year={2002}
}

@inproceedings{hassan2012obebangladesh,
  title={Challenges of implementing outcome based engineering education in universities in Bangladesh},
  author={Hassan, MM Shahidul},
  booktitle={Proceedings of the 7th International Conference on Electrical and Computer Engineering (ICECE)},
  pages={362--364},
  year={2012},
  doi={10.1109/ICECE.2012.6471562}
}

@article{hu2021lora,
  title={{LoRA}: Low-rank adaptation of large language models},
  author={Hu, Edward J and Shen, Yelong and Wallis, Phillip and Allen-Zhu, Zeyuan and Li, Yuanzhi and Wang, Shean and Wang, Lu and Chen, Weizhu},
  journal={arXiv preprint arXiv:2106.09685},
  year={2021}
}

@article{jonsson2007rubrics,
  title={The use of scoring rubrics: Reliability, validity and educational consequences},
  author={Jonsson, Anders and Svingby, Gunilla},
  journal={Educational research review},
  volume={2},
  number={2},
  pages={130--144},
  year={2007},
  publisher={Elsevier}
}

@article{khosravi2022explainable,
  title={Explainable artificial intelligence in education},
  author={Khosravi, Hassan and Shum, Simon Buckingham and Chen, Guanliang and Conati, Cristina and Tsai, Yi-Shan and Kay, Judy and Knight, Simon and Martinez-Maldonado, Roberto and Sadiq, Shazia and Ga{\v{s}}evi{\'c}, Dragan},
  journal={Computers and Education: Artificial Intelligence},
  volume={3},
  pages={100074},
  year={2022}
}

@inproceedings{kim2022donut,
  title={{OCR}-free document understanding transformer},
  author={Kim, Geewook and Hong, Teakgyu and Yim, Moonbin and Nam, JeongYeon and Park, Jinyoung and Yim, Jinyeong and Hwang, Wonseok and Yun, Sangdoo and Han, Dongyoon and Park, Seunghyun},
  booktitle={Computer Vision -- ECCV 2022},
  pages={498--517},
  year={2022},
  doi={10.1007/978-3-031-19815-1_29}
}

@article{landis1977measurement,
  title={The measurement of observer agreement for categorical data},
  author={Landis, J Richard and Koch, Gary G},
  journal={Biometrics},
  volume={33},
  number={1},
  pages={159--174},
  year={1977}
}

@inproceedings{li2023trocr,
  title={{TrOCR}: Transformer-based optical character recognition with pre-trained models},
  author={Li, Minghao and Lv, Tengchao and Chen, Jingye and Cui, Lei and Lu, Yijuan and Florencio, Dinei and Zhang, Cha and Li, Zhoujun and Wei, Furu},
  booktitle={Proceedings of the AAAI Conference on Artificial Intelligence},
  volume={37},
  number={11},
  pages={13094--13102},
  year={2023}
}

@inproceedings{li2022blip,
  title={{BLIP}: Bootstrapping language--image pre-training for unified vision--language understanding and generation},
  author={Li, Junnan and Li, Dongxu and Xiong, Caiming and Hoi, Steven},
  booktitle={Proceedings of the 39th International Conference on Machine Learning},
  volume={162},
  pages={12888--12900},
  year={2022}
}

@article{liu2023llava,
  title={Visual instruction tuning},
  author={Liu, Haotian and Li, Chunyuan and Wu, Qingyang and Lee, Yong Jae},
  journal={Advances in Neural Information Processing Systems},
  volume={36},
  year={2023}
}

@inproceedings{ma2024mmlongbench,
  title={{MMLongBench-Doc}: Benchmarking long-context document understanding with visualizations},
  author={Ma, Yubo and Zang, Yuhang and Chen, Liangyu and Chen, Meiqi and Jiao, Yizhu and Li, Xinze and Lu, Xinyuan and Liu, Ziyu and Ma, Yan and Dong, Xiaoyi and Zhang, Pan and Pan, Liangming and Jiang, Yu-Gang and Wang, Jiaqi and Cao, Yixin and Sun, Aixin},
  booktitle={Advances in Neural Information Processing Systems 37: Datasets and Benchmarks Track},
  year={2024}
}

@misc{mangrulkar2022peft,
  title={{PEFT}: State-of-the-art parameter-efficient fine-tuning methods},
  author={Mangrulkar, Sourab and Gugger, Sylvain and Debut, Lysandre and Belkada, Younes and Paul, Sayak and Bossan, Benjamin and Tietz, Marian},
  howpublished={\url{https://github.com/huggingface/peft}},
  year={2022}
}

@article{mistral2024pixtral,
  title={Pixtral 12B},
  author={Agrawal, Pravesh and Antoniak, Szymon and Hanna, Emma Bou and Bout, Baptiste and Chaplot, Devendra and Chudnovsky, Jessica and Costa, Diogo and De Monicault, Baudouin and Garg, Saurabh and Gervet, Theophile and others},
  journal={arXiv preprint arXiv:2410.07073},
  year={2024}
}

@inproceedings{mondal2024icdarhwd,
  title={{ICDAR} 2024 competition on recognition and {VQA} on handwritten documents},
  author={Mondal, Ajoy and Mahadevan, Vijay and Manmatha, R. and Jawahar, C. V.},
  booktitle={Document Analysis and Recognition -- ICDAR 2024},
  year={2024},
  doi={10.1007/978-3-031-70552-6_26}
}

@article{page1968analysis,
  title={The use of the computer in analyzing student essays},
  author={Page, Ellis B},
  journal={International review of education},
  pages={210--225},
  year={1968},
  publisher={JSTOR}
}

@inproceedings{paszke2019pytorch,
  title={{PyTorch}: An imperative style, high-performance deep learning library},
  author={Paszke, Adam and Gross, Sam and Massa, Francisco and others},
  booktitle={Advances in Neural Information Processing Systems},
  volume={32},
  pages={8024--8035},
  year={2019}
}

@article{premalatha2019course,
      title={Course and program outcomes assessment methods in outcome-based education: A review},
  author={Premalatha, Kandhasamy},
  journal={Journal of Education},
  volume={199},
  number={3},
  pages={111--127},
  year={2019}
}

@inproceedings{radford2021clip,
  title={Learning transferable visual models from natural language supervision},
  author={Radford, Alec and Kim, Jong Wook and Hallacy, Chris and Ramesh, Aditya and Goh, Gabriel and Agarwal, Sandhini and Sastry, Girish and Askell, Amanda and Mishkin, Pamela and Clark, Jack and others},
  booktitle={Proceedings of the 38th International Conference on Machine Learning},
  year={2021}
}

@article{rajak2019copo,
  title={An approach to evaluate program outcomes and program educational objectives through direct and indirect assessment tools},
  author={Rajak, Akash and Shrivastava, Ajay Kumar and Tripathi, Arun Kumar},
  journal={International Journal of Emerging Technologies in Learning},
  volume={14},
  number={23},
  pages={85--97},
  year={2019}
}

@inproceedings{ramchandra2014attainment,
  title={Method for estimation of attainment of program outcome through course outcome for outcome based education},
  author={Ramchandra, Shivakumar and Maitra, Samita and MallikarjunaBabu, K},
  booktitle={2014 IEEE International Conference on MOOC, Innovation and Technology in Education (MITE)},
  pages={7--12},
  year={2014},
  organization={IEEE},
  doi={10.1109/MITE.2014.7020231}
}

@article{ramesh2022autonomous,
  title={An automated essay scoring systems: A systematic literature review},
  author={Ramesh, Dadi and Sanampudi, Suresh Kumar},
  journal={Artificial Intelligence Review},
  volume={55},
  number={3},
  pages={2495--2527},
  year={2022}
}

@article{reddy2021copomapping,
  title={A case study on the assessment of program quality through {CO--PO} mapping and its attainment},
  author={Rajagopal Reddy, B. and Karuppiah, N. and Asif, M. and Ravivarman, S.},
  journal={Journal of Engineering Education Transformations},
  volume={34},
  pages={104--111},
  year={2021},
  doi={10.16920/jeet/2021/v34i0/157114}
}

@inproceedings{ridley2021automated,
  title={Automated cross-prompt scoring of essay traits},
  author={Ridley, Robert and He, Liang and Dai, Xinyu and Huang, Shujian and Chen, Jiajun},
  booktitle={Proceedings of the AAAI Conference on Artificial Intelligence},
  volume={35},
  number={15},
  pages={13745--13753},
  year={2021}
}

@article{rodrigues2025gpt4fair,
  title={Is {GPT-4} fair? An empirical analysis in automatic short answer grading},
  author={Rodrigues, Luiz and Xavier, Cleon and Costa, Newarney and Ga{\v{s}}evi{\'c}, Dragan and Mello, Rafael Ferreira},
  journal={Computers and Education: Artificial Intelligence},
  volume={8},
  year={2025},
  doi={10.1016/j.caeai.2025.100428}
}

@book{shermis2003automated,
  title={Automated essay scoring: A cross-disciplinary perspective},
  author={Shermis, Mark D and Burstein, Jill C},
  year={2003},
  publisher={Routledge}
}

@book{shermis2013handbook,
  title={Handbook of automated essay evaluation: Current applications and new directions},
  author={Shermis, Mark D and Burstein, Jill},
  year={2013},
  publisher={Routledge}
}

@book{spady1994outcome,
  title={Outcome-Based Education: Critical Issues and Answers},
  author={Spady, William G},
  publisher={American Association of School Administrators},
  address={Arlington, VA, USA},
  year={1994}
}

@inproceedings{taghipour2016neural,
  title={A neural approach to automated essay scoring},
  author={Taghipour, Kaveh and Ng, Hwee Tou},
  booktitle={Proceedings of the 2016 conference on empirical methods in natural language processing},
  pages={1882--1891},
  year={2016}
}

@article{tisi2013review,
  title={A review of literature on marking reliability research},
  author={Tisi, Jo and Whitehouse, Gillian and Maughan, Sarah and Burdett, Newman},
  journal={Ofqual/13/5285},
  year={2013}
}

@article{williamson2012framework,
  title={A framework for evaluation and use of automated scoring},
  author={Williamson, David M and Xi, Xiaoming and Breyer, F Jay},
  journal={Educational measurement: issues and practice},
  volume={31},
  number={1},
  pages={2--13},
  year={2012},
  publisher={Wiley Online Library}
}

@inproceedings{wolf2020transformers,
  title={Transformers: State-of-the-art natural language processing},
  author={Wolf, Thomas and Debut, Lysandre and Sanh, Victor and Chaumond, Julien and Delangue, Clement and Moi, Anthony and Cistac, Pierric and Rault, Tim and Louf, R{\'e}mi and Funtowicz, Morgan and others},
  booktitle={Proceedings of the 2020 conference on empirical methods in natural language processing: system demonstrations},
  pages={38--45},
  year={2020}
}

@inproceedings{wu2024faithfulness,
  title={On the faithfulness of vision transformer explanations},
  author={Wu, Junyi and Kang, Weitai and Tang, Hao and Hong, Yuan and Yan, Yan},
  booktitle={2024 IEEE/CVF Conference on Computer Vision and Pattern Recognition (CVPR)},
  year={2024},
  organization={IEEE}
}

@article{zhu2025internvl3,
  title={{InternVL3}: Exploring advanced training and test-time recipes for open-source multimodal models},
  author={Zhu, Jinguo and Wang, Weiyun and Chen, Zhe and Liu, Zhaoyang and Ye, Shenglong and Gu, Lixin and Tian, Hao and Duan, Yuchen and Su, Weijie and Shao, Jie and others},
  journal={arXiv preprint arXiv:2504.10479},
  year={2025}
}

\newcommand{\authorphoto}[1]{%
  \IfFileExists{#1}{%
    \includegraphics[width=1in,height=1.25in,clip,keepaspectratio]{#1}%
  }{%
    \fbox{\parbox[c][1.15in][c]{0.88in}{%
      \centering\footnotesize PLACE\\PHOTO\\HERE}}%
  }%
}

\begin{IEEEbiography}[{\authorphoto{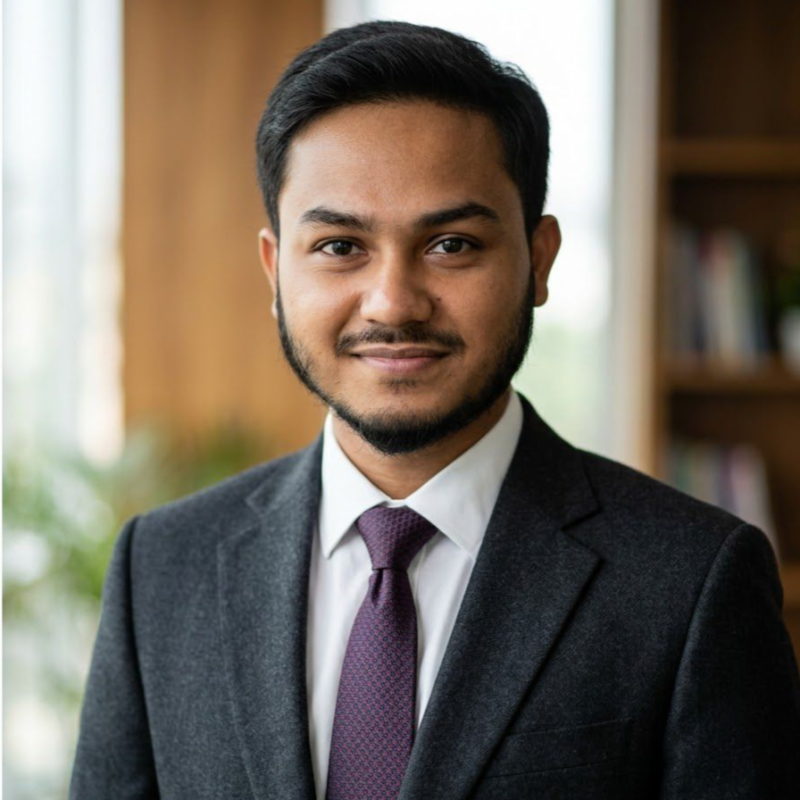}}]{Asif Hasan Tonmoy} received the B.Sc. degree in Computer Science and Engineering from Khulna University of Engineering and Technology (KUET), Khulna, Bangladesh, in 2023. He is a Full-Stack Software Engineer and technology consultant with experience delivering software and AI solutions, technical consultancy, and digital growth strategies for national and international clients.

His areas of expertise include large language models, retrieval-augmented generation, agentic AI, semantic search, vector databases, full-stack development, scalable software architecture, cloud infrastructure, and production software solutions. His interests focus on designing reliable and scalable AI-powered software systems that align technical solutions with practical business requirements.

\end{IEEEbiography}

\begin{IEEEbiography}[{\authorphoto{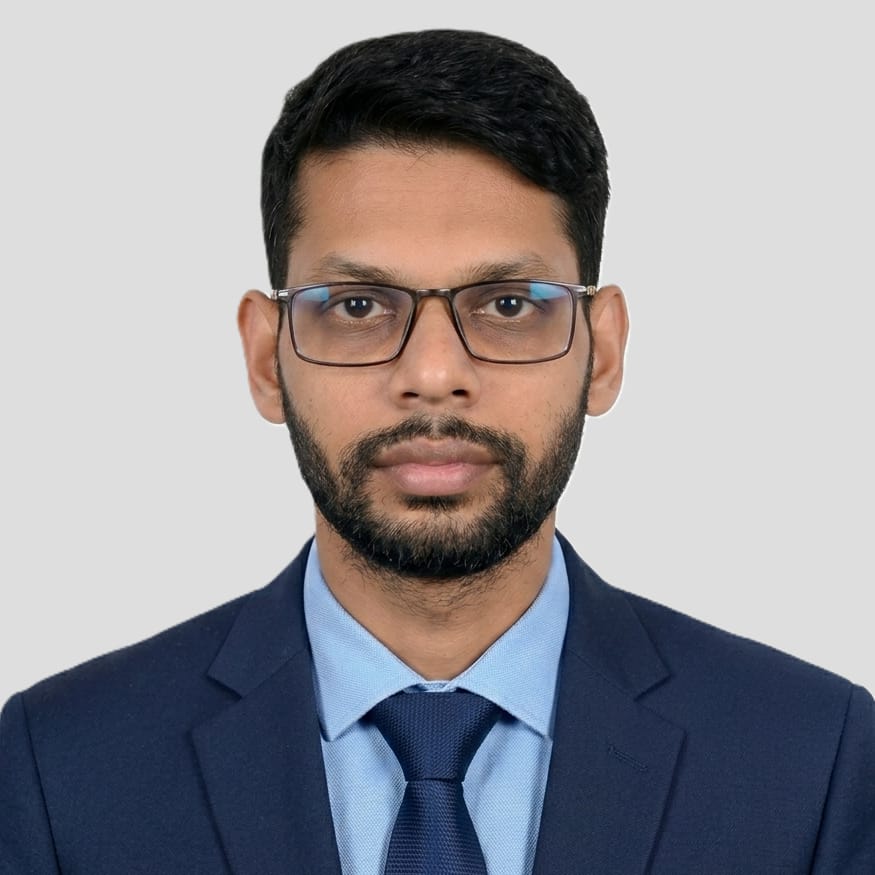}}]{Saad Ahmed} received the B.Sc. degree in Computer Science and Engineering from Bangladesh Army University of Science and Technology (BAUST), Saidpur, Bangladesh, in 2024. He is currently pursuing the M.Sc. degree in Computer Science and Engineering at Rajshahi University of Engineering and Technology (RUET), Rajshahi, Bangladesh.

Since 2024, he has been serving as a Lecturer with the Department of Computer Science and Engineering, Bangladesh Army University of Science and Technology (BAUST), Saidpur, Bangladesh.

His research interests include computer vision, deep learning, explainable artificial intelligence, and multimodal learning architectures, with particular applications in agricultural image analysis and document image understanding.
\end{IEEEbiography}

\begin{IEEEbiography}[{\authorphoto{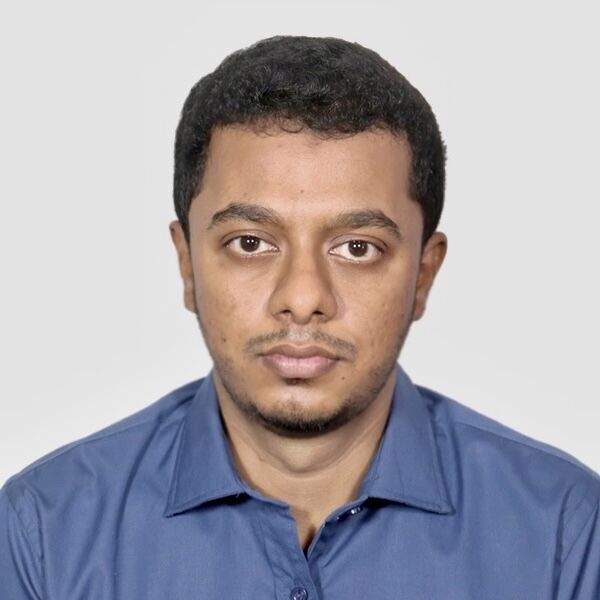}}]{Md Khalid Syfullah}
(Member, IEEE) received the B.Sc. degree in Computer Science and Engineering from Khulna University of Engineering and Technology (KUET), Khulna, Bangladesh, in 2023. He is currently pursuing the Ph.D. degree in Computer Science at Southern Illinois University Carbondale, Carbondale, IL, USA. Before beginning his doctoral studies, he was involved in teaching, research, and mentoring activities in computer science.

His research interests include trustworthy and privacy-preserving artificial intelligence, distributed and federated learning, and vision-language models. His current research focuses on improving the security and privacy of vision-language models.
\end{IEEEbiography}

\begin{IEEEbiography}[{\authorphoto{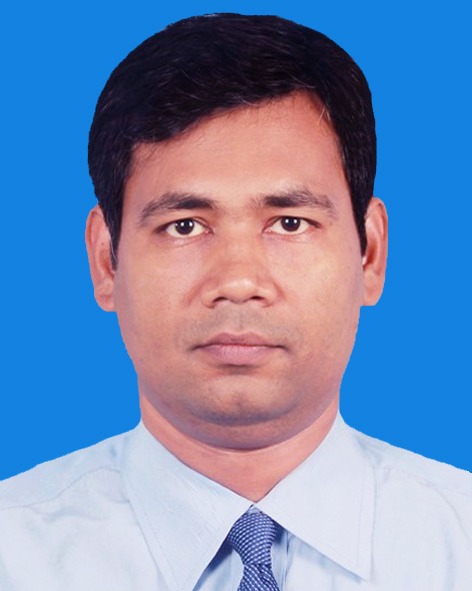}}]{S. M. Jahangir Alam} received the Ph.D. degree in Mechatronic Engineering, specializing in artificial intelligence and computer vision, from Xiamen University, Xiamen, China, in 2014. He has held several academic and research positions, including Associate Professor with the South China University of Technology, China, and visiting positions at the University of British Columbia, Canada; International Islamic University Malaysia (IIUM); Universiti Tun Hussein Onn Malaysia (UTHM); and Xiamen Ocean University, China. Since February 2022, he has been a Professor with the Department of Computer Science and Engineering, Bangladesh Army University of Science and Technology (BAUST), Bangladesh.

He holds seven patents and has authored or coauthored more than 38 research publications, including 16 SCI/EI-indexed articles. He has contributed to 19 research projects as a principal investigator or participant and serves as a reviewer and editorial board member for scholarly journals. He has also authored five reference and textbooks covering machine learning, computer vision, deep learning, data science, and control engineering.

His research interests include artificial intelligence, data science, natural language processing, computer vision, machine learning, deep learning, the Internet of Things, robotics, healthcare and mobile health, biomedical engineering, and intelligent systems.

\end{IEEEbiography}

\end{document}